%% file: main.tex
\pdfoutput=1

\documentclass[11pt]{article}

\usepackage[final]{acl}

\usepackage{times}
\usepackage{latexsym}
\usepackage{float}
\usepackage{placeins}

\usepackage[T1]{fontenc}

\usepackage[utf8]{inputenc}

\usepackage{microtype}

\usepackage{inconsolata}
\usepackage{subcaption}
\usepackage{graphicx}
\usepackage{enumitem}
\usepackage{amsmath}
\usepackage{url}            
\usepackage{booktabs}       
\usepackage{amsfonts}       
\usepackage{nicefrac}       
\usepackage{microtype}      
\usepackage{CJKutf8}
\usepackage{parskip}
\usepackage{colortbl}
\usepackage{graphicx} 
\usepackage[dvipsnames]{xcolor} 
\usepackage{pifont} 
\usepackage{multirow}
\usepackage{lscape}
\definecolor{mygray}{gray}{.88}

\newlength\savewidth

\usepackage{parskip}
\usepackage{algorithm}
\usepackage{algpseudocode}
\usepackage{listings}
\definecolor{pyblue}{rgb}{0.0, 0.0, 0.5}
\definecolor{pygreen}{rgb}{0.0, 0.5, 0.0}
\definecolor{pyorange}{rgb}{1.0, 0.4, 0.0}

\usepackage{tikz} 

\newcommand{\solidcircle}[1]{%
  \tikz[baseline=(char.base)]{
    \node[shape=circle, draw, fill=black, text=white, inner sep=1pt] (char) {\textbf{\small #1}};
  }%
}

\lstdefinestyle{pythonstyle}{
    language=Python,
    basicstyle=\footnotesize\ttfamily,
    breaklines=true,
    morekeywords={self},
    keywordstyle=\color{pyblue},
    commentstyle=\color{pygreen},
    stringstyle=\color{pyorange},
    numberstyle=\tiny\color{gray},
    numbers=left,
    numbersep=10pt,
    tabsize=4,
    showspaces=false,
    showstringspaces=false
}

\usepackage{tcolorbox}
\usepackage{fontawesome5}

\usepackage{alltt}
\newcommand{\shortname}{\textsc{Amanda}}
\title{\shortname{}: Agentic Medical Knowledge Augmentation for Data-Efficient Medical Visual Question Answering}

\author{
 \textbf{Ziqing Wang\textsuperscript{1}}, 
 \textbf{Chengsheng Mao\textsuperscript{1}}, 
 \textbf{Xiaole Wen\textsuperscript{2}},  \\
 \textbf{Yuan Luo\textsuperscript{1}}\thanks{Co-corresponding authors.}, 
 \textbf{Kaize Ding\textsuperscript{1}}\footnotemark[1]
\\
 \textsuperscript{1}Northwestern University, 
 \textsuperscript{2}Microsoft
}

\begin{document}
\maketitle
\input{sec/0_abstract}

\input{sec/1_intro}

\input{sec/2_related_work}

\input{sec/3_method}

\input{sec/4_experiments}

\input{sec/5_conclusion}
\input{sec/6_limitations}
\bibliography{main}

\appendix
\clearpage
\newpage

\input{sec/X_suppl}

\end{document}

%% file: sec/0_abstract.tex
\begin{abstract} Medical Multimodal Large Language Models (Med-MLLMs) have shown great promise in medical visual question answering (Med-VQA). However, when deployed in low-resource settings where abundant labeled data are unavailable, existing Med-MLLMs commonly fail due to their medical reasoning capability bottlenecks: (i) the intrinsic reasoning bottleneck that ignores the details from the medical image; (ii) the extrinsic reasoning bottleneck that fails to incorporate specialized medical knowledge. To address those limitations, we propose \textbf{\shortname{}}, a training-free agentic framework that performs medical knowledge augmentation via LLM agents. Specifically, our intrinsic medical knowledge augmentation focuses on coarse-to-fine question decomposition for comprehensive diagnosis, while extrinsic medical knowledge augmentation grounds the reasoning process via biomedical knowledge graph retrieval. Extensive experiments across eight Med-VQA benchmarks demonstrate substantial improvements in both zero-shot and few-shot Med-VQA settings. The code is available at \url{https://github.com/REAL-Lab-NU/AMANDA}.
\end{abstract}



%% file: sec/1_intro.tex
\section{Introduction} \label{sec: intro}


Medical Visual Question Answering (Med-VQA) aims to automatically answer natural language questions about medical images, serving as an AI-powered assistant to enhance healthcare professionals' diagnostic efficiency and accuracy~\cite{hartsock2024vision, lin2023medical}. Unlike general-domain VQA which focuses on everyday scenes and objects, Med-VQA requires fine-grained analysis of subtle pathological features, understanding of professional medical terminology, and integration of domain-specific medical knowledge~\cite{lin2023medical}. These unique characteristics make Med-VQA particularly challenging yet crucial for empowering precise medical diagnosis.

Recent advances in Medical Multimodal Large Language Models (Med-MLLMs) have demonstrated promising results in Med-VQA through extensive pre-training and task-specific fine-tuning~\cite{li2024llava, eslami2023pubmedclip, zhang2023large, jiang2024medmoemixturedomainspecificexperts}. However, obtaining a large-scale medical dataset for Med-MLLM pre-training or fine-tuning requires labor-intensive expert annotations, making it impractical in data-efficient scenarios. When deployed in low-resource settings where abundant training or fine-tuning data are unavailable (i.e., zero-shot or few-shot settings), existing Med-MLLMs commonly fail due to two bottlenecks of their medical reasoning capability:
\begin{itemize}[leftmargin=*, itemsep=0.0pt, topsep=0pt]
    \item From the \textit{intrinsic} perspective, current Med-MLLMs usually focus on understanding the image from a general view, while ignoring the fine-grained examination of subtle pathological features that are critical for accurate diagnosis~\cite{lin2023medical}. In clinical practice, medical professionals achieve comprehensive analysis through an iterative process of questioning and examination, progressively uncovering crucial details. However, the single-step inference adopted by existing Med-MLLMs fails to capture this iterative nature of the medical diagnosis, leading to superficial analyses without critical diagnostic details~\cite{wang2023towards, jiang2024multia, jiang2024multib}.
    \item From the \textit{extrinsic} perspective, while Med-MLLMs possess basic medical knowledge through pre-training, these models are typically static and lack mechanisms to access or incorporate new medical knowledge continually. In Med-VQA tasks, such specialized medical knowledge from up-to-date knowledge bases is particularly crucial. Correspondingly, existing methods often struggle to provide comprehensive and contextually grounded answers, with a concerning tendency to generate hallucinations~\cite{xia2024rule, yan2024worse} -- plausible but factually incorrect responses that pose significant risks for real-world medical diagnosis. 
\end{itemize}

To address the aforementioned challenges, we present a training-free MLLM agentic framework -- \shortname{} (\underline{A}gentic \underline{M}edic\underline{A}l K\underline{N}owle\underline{D}ge \underline{A}ugmentation) for data-efficient medical visual question answering. In essence, our framework enhances Med-MLLMs' reasoning capability through \textit{Medical Knowledge Augmentation} (Med-KA) from both intrinsic and extrinsic reasoning perspectives. On the one hand, to enhance the medical reasoning depth, we propose \textit{Intrinsic Med-KA}, which leverages a coarse-to-fine question decomposition strategy to fully utilize the intrinsic visual understanding capabilities within Med-MLLMs, enabling comprehensive diagnosis through progressive examination. On the other hand, to bridge the gap between models' pre-trained knowledge and reliable medical expertise, we develop \textit{Extrinsic Med-KA}, which retrieves relevant medical knowledge from biomedical knowledge graphs to ground the reasoning process. These complementary approaches are orchestrated by multiple LLM agents that can adaptively control the depth of knowledge integration to maintain both effectiveness and efficiency. In addition, \shortname{} can incorporate in-context learning examples, enabling further performance gains in few-shot settings. Overall, our contributions can be summarized as follows:
\vspace{-0.2cm}
\begin{itemize}[leftmargin=*, itemsep=0pt, topsep=0pt]
\item \textbf{Problem.} We target the challenging problem of data-efficient Med-VQA and propose a training-free agentic framework that addresses the intrinsic and extrinsic bottlenecks of Med-MLLMs' reasoning capability via Med-KA.
\vspace{-0.2cm}
\item \textbf{Method.} We develop a Med-KA approach from two complementary perspectives: \textit{intrinsic Med-KA} through coarse-to-fine question decomposition and \textit{extrinsic Med-KA} via medical knowledge graph retrieval, unified under an adaptive refinement mechanism.
\vspace{-0.2cm}
\item \textbf{Experiments.} Through comprehensive experiments on eight Med-VQA benchmarks, we demonstrate substantial improvements in both zero-shot and few-shot settings, with strong generalization across different types of MLLMs.
\end{itemize}

%% file: sec/2_related_work.tex
\section{Related Work} \label{sec: related_work}

\paragraph{Medical Visual Question Answering.} Current Med-VQA approaches primarily follow two paradigms: discriminative methods that select from predefined options~\cite{zhang2023large, eslami2023pubmedclip}, and generative methods that enable open-ended responses~\cite{bazi2023vision, liu2023q2atransformer, van2023open, wei2024mc}. While discriminative methods achieve high performance in controlled settings, their predefined answer space limits applicability in real-world medical scenarios. Recent Med-MLLMs~\cite{li2024llava, jiang2024medmoemixturedomainspecificexperts} have shown promising results with flexible response generation. However, they require extensive labeled data for training and fine-tuning. To address this limitation, our \shortname{} introduces a novel MLLM agentic framework for data-efficient scenarios without task-specific fine-tuning.

\begin{figure*}[t!]
\centering
\includegraphics[height=3.8cm]{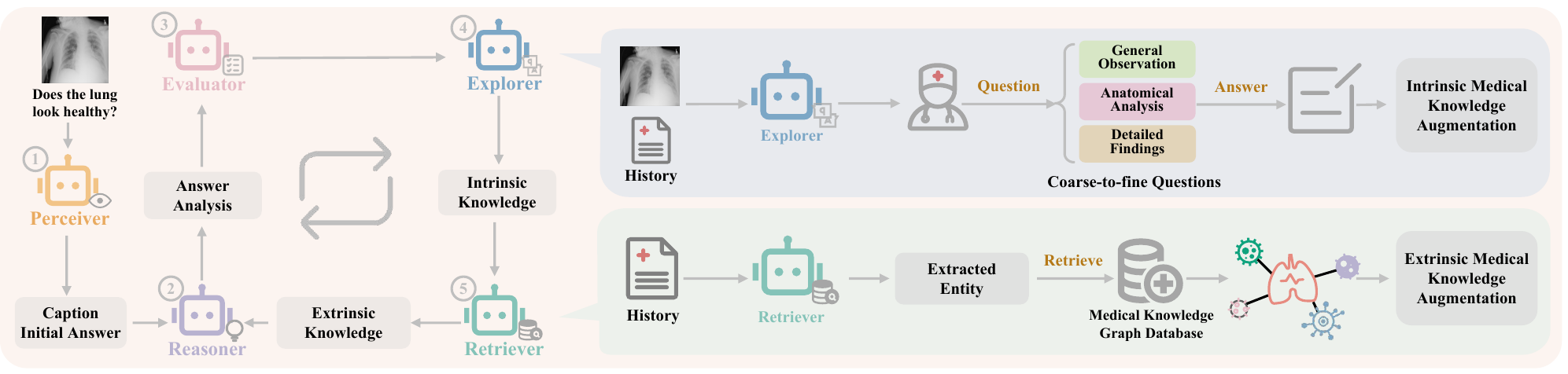}
\caption{\textbf{Overview of our \shortname{} framework.} The framework comprises five specialized agents (\texttt{Perceiver}, \texttt{Reasoner}, \texttt{Evaluator}, \texttt{Explorer}, and \texttt{Retriever}) working collaboratively to enable comprehensive and reliable medical reasoning. Specifically, the \texttt{Explorer} incorporates intrinsic medical knowledge through coarse-to-fine question decomposition to enhance reasoning depth, and the \texttt{Retriever} integrates extrinsic medical knowledge from biomedical knowledge graphs to enable reliable medical reasoning. The \texttt{Evaluator} adaptively controls the depth of Med-KA to enable efficient and accurate medical diagnosis.} 
\label{fig: pipeline}
\vspace{-0.3cm}
\end{figure*}

\paragraph{Large Multimodal Agent.} Recent research has demonstrated the effectiveness of combining LLMs' reasoning capabilities~\cite{chatgpt, gpt4} with MLLMs for visual tasks. Early works like PNP-VQA~\cite{tiong2022plug} and Img2LLM~\cite{guo2023images} demonstrated the effectiveness of integrating visual understanding with LLMs' reasoning capabilities. This integration has evolved into sophisticated large multimodal agent systems~\cite{you2023idealgpt, suris2023vipergpt, wu2023visual, xie2024large}, where multiple LLM-powered agents collaborate. However, in the medical domain, most existing agent systems~\cite{tang2023medagents, fan2024ai, schmidgall2024agentclinic, wei2024medco, li2024agent, kim2024adaptive} primarily focus on text-based scenarios, lacking crucial multimodal capabilities. While recent work like MMedAgent~\cite{li2024mmedagent} explores multimodal agents for medical applications, it requires extensive task-specific training, limiting its applicability in data-efficient settings. Our \shortname{} addresses these limitations by introducing a training-free MLLM agentic framework for data-efficient medical visual reasoning. 

\paragraph{Medical Knowledge Augmentation.} Integrating medical knowledge has proven essential for enhancing medical AI systems~\cite{fang2019attention, gonzalez2018dermaknet, wang2020learning, chen2022align, tan2019expert, chen2020generating, soman2023biomedical, wu2023medklip, nath2025vila}. Representative works include Med-VLP~\cite{chen2022align}, which employs UMLS Knowledge Graph~\cite{bodenreider2004unified} for cross-modal alignment, and KG-RAG~\cite{soman2023biomedical}, which leverages biomedical knowledge graphs with LLMs. Building upon these advances, our \shortname{} introduces a holistic knowledge augmentation approach to enable comprehensive and reliable medical reasoning.

%% file: sec/3_method.tex
\section{Proposed Approach -- \shortname{}} \label{sec: method}
While existing Med-MLLMs have shown promise, they often struggle in data-efficient settings due to critical reasoning and knowledge bottlenecks. To address these limitations, we introduce AMANDA, a training-free, agentic framework specifically designed for Med-VQA. In this section, we first formalize the Med-VQA problem and present our \shortname{} framework (Sec.~\ref{sec: formulation} and \ref{sec: architecture}). We then detail our Med-KA approaches (Sec.~\ref{sec: aka}) and present two extensions: the adaptive reasoning refinement mechanism and the few-shot enhancement strategies (Sec.~\ref{sec: extension}).

\subsection{Problem Formulation} \label{sec: formulation}
We target Med-VQA in data-efficient scenarios, particularly zero-shot and few-shot settings, where task-specific training data is limited or unavailable. Traditional Med-VQA approaches~\cite{li2024llava, eslami2023pubmedclip, zhang2023large} typically employ a single Med-MLLM for direct inference. Following previous works~\cite{zhang2023pmc}, this process can be formulated as:
$$
\hat{a}=\Phi_{\text{MedVQA}}\left(\mathcal{I}, q\right)
$$
where $\hat{a}$ is the output answer, $\mathcal{I} \in \mathbb{R}^{H \times W \times C}$ represents the input medical image with height $H$, width $W$, and channel number $C$, $q$ denotes the question, and $\Phi$ is the Med-MLLM model.

However, this single-step approach, directly adapted from the general domain, faces two critical limitations in medical image analysis~\cite{liu2024visual}. First, it fails to systematically examine multiple aspects of medical images, often missing subtle details that are crucial for differentiating similar conditions—a limitation we term the intrinsic reasoning bottleneck. Second, in data-efficient scenarios where models encounter novel cases, the lack of comprehensive medical knowledge leads to unreliable analysis or hallucinations~\cite{xia2024rule, yan2024worse}—what we identify as the extrinsic reasoning bottleneck.. 

To address these limitations, we reformulate Med-VQA as an iterative reasoning process that leverages multiple specialized agents:
\[ \hat{a}_t = \Phi_{\text{iterative}}(\mathcal{I}, q, \mathcal{H}_{t-1} \cup \bigcup_{i \in \mathcal{A}} h_t^i) \] 
where $\hat{a}_t$ represents the refined answer at iteration $t$, $\Phi_{\text{iterative}}$ denotes our proposed iterative reasoning framework, $\mathcal{H}_{t-1}$ is the accumulated reasoning history up to iteration $t-1$, $\mathcal{A}$ represents our agent set and $h_t^i$ denotes each agent's output at iteration $t$. This formulation transforms the single-step approach into an iterative reasoning process where specialized agents collaboratively refine the answer through progressive analysis.

\subsection{Architecture Overview} \label{sec: architecture}
To enable such iterative medical reasoning, we design an agentic framework -- \shortname{}. Our framework comprises three functional modules, where specialized agents work collaboratively:

\begin{itemize} [leftmargin=*, itemsep=0.0pt]
    \item \noindent\textbf{Perception Module.} The \texttt{Perceiver} agent, implemented using a Med-MLLM (e.g., LLaVA-Med v1.5~\cite{li2024llava}), establishes the foundation for visual analysis. Unlike single-step approaches~\cite{li2024llava} that directly generate answers, our \texttt{Perceiver} provides two outputs: a detailed medical caption $c$ and an initial answer $\hat{a}_0$ to the main question. The medical caption $c$ is generated through carefully designed prompts (see Appendix~\ref{appendix: prompt}) to systematically describe general observation. The initial answer $\hat{a}_0$, while potentially imperfect, provides a basic foundation that will be progressively refined. Together, these outputs enable more accurate and comprehensive analysis in subsequent modules.

    \item \noindent\textbf{Planning Module.} Building upon the Perception Module's outputs, the Planning Module coordinates the overall reasoning process through two LLM-based agents. The \texttt{Reasoner} analyzes the available information (medical caption, initial answer, and any augmented knowledge) to generate a refined answer through systematic medical reasoning. The \texttt{Evaluator} then assesses the reasoning quality through a confidence score, determining whether additional knowledge augmentation is needed (detailed in Sec.~\ref{sec: extension}). This grants the framework key agentic properties like autonomous decision-making and adaptive behavior, moving beyond a static, pre-defined pipeline.

    \item \noindent\textbf{Action Module.} Triggered by the Planning Module, the Action Module addresses both reasoning bottlenecks through two complementary knowledge augmentation agents. From the intrinsic perspective, the \texttt{Explorer}, powered by LLM, enhances the visual reasoning depth by decomposing the original question $q$ into sub-questions $q_{sub}$, which are then answered by the same Med-MLLM used in the \texttt{Perceiver}. From the extrinsic perspective, the \texttt{Retriever}, also implemented using LLM, grounds the analysis by retrieving and integrating relevant medical knowledge from biomedical knowledge graphs. Both agents' outputs are fed back to the Planning Module for further answer refinement.
\end{itemize}

\noindent\textbf{Collaborative Medical Reasoning Workflow.}
Our \shortname{} framework orchestrates these three modules in a collaborative workflow. As shown in Fig.~\ref{fig: pipeline}: \solidcircle{1} The \texttt{Perceiver} performs visual analysis to generate a general medical caption and an initial answer. \solidcircle{2} The \texttt{Reasoner} synthesizes all the available information to produce a refined answer. \solidcircle{3} The \texttt{Evaluator} assesses the confidence of current answer. \solidcircle{4} When additional knowledge is needed, the \texttt{Explorer} and \texttt{Retriever} performs both intrinsic Med-KA and extrinsic Med-KA. This augmented knowledge is then fed back to the \texttt{Reasoner} for further refinement. Communication among agents is managed through a centralized reasoning history, where each agent's output is appended, allowing subsequent agents to build upon a coherent and progressively enriched context. This iterative process inherently enables a form of self-correction, as the \texttt{Reasoner} can revise the initial answer based on newly acquired visual details from the \texttt{Explorer} or grounded facts from the \texttt{Retriever}.

\subsection{Medical Knowledge Augmentation with LLM Agents} \label{sec: aka}
Building upon our agentic framework, we now detail our medical knowledge augmentation strategies that enhance Med-MLLMs' reasoning capability in data-efficient scenarios.

\noindent\textbf{Intrinsic Medical Knowledge Augmentation.}
In data-efficient scenarios where abundant training data is unavailable, Med-MLLMs often struggle with comprehensive visual analysis due to their single-step inference approach. For instance, when asked \textit{"Does the chest X-ray look healthy?"}, models typically provide general responses like \textit{"no obvious abnormalities"} without examining key diagnostic features. This limitation stems from the lack of progressive questioning in single-step inference, where models fail to focus on specific yet crucial details, resulting in superficial responses that overlook critical diagnostic features.

To address this intrinsic bottleneck, we draw inspiration from the question decomposition strategy, where complex problems are broken down into focused sub-questions for comprehensive analysis. Recent studies have demonstrated that LLMs possess strong capabilities in reasoning enhancement through question decomposition~\cite{wu2023visual, suris2023vipergpt, zhu2023chatgpt, you2023idealgpt}. These methods leverage LLMs to decompose complex tasks into manageable sub-questions, enabling progressive understanding through structured questioning. Motivated by these advances, we adapt this approach to medical visual analysis to enable deeper and more thorough reasoning. Specifically, we propose a coarse-to-fine intrinsic Med-KA strategy through our \texttt{Explorer} agent.

The strategy is triggered when the \texttt{Evaluator} detects insufficient reasoning depth in the \texttt{Reasoner}'s analysis. Our \texttt{Explorer} agent consists of two key components: (1) an LLM-powered questioning component that analyzes the main question, medical caption, and current reasoning history to generate targeted follow-up questions, and (2) an answering component that utilizes the same Med-MLLM as in the \texttt{Perceiver} to provide detailed analysis for each question. At each iteration, \texttt{Explorer} generates three follow-up questions and their corresponding answers in a hierarchical strategy:
\begin{itemize}[leftmargin=3mm]\setlength{\itemsep}{-5pt}
    \item \textbf{General Observation.} First focuses on overall appearance and key findings (e.g., \textit{``What is the overall appearance of the image?"}), establishing a foundation for medical analysis.
    \item \textbf{Anatomical Analysis.} Then examines specific anatomical regions or structures, considering their characteristics (size, shape, alignment) and spatial relationships (e.g., \textit{``What is the appearance and position of the cardiac silhouette?"}). 
    \item \textbf{Detailed Findings.} Finally investigates potential pathological features in regions of interest (e.g., \textit{"Are there any infiltrates or masses in the lower right lung field, and what are their specific characteristics?"}), enabling the detection of subtle abnormalities through focused analysis.
\end{itemize}

\begin{figure*}[t!]
\centering  
\includegraphics[height=4.3cm]{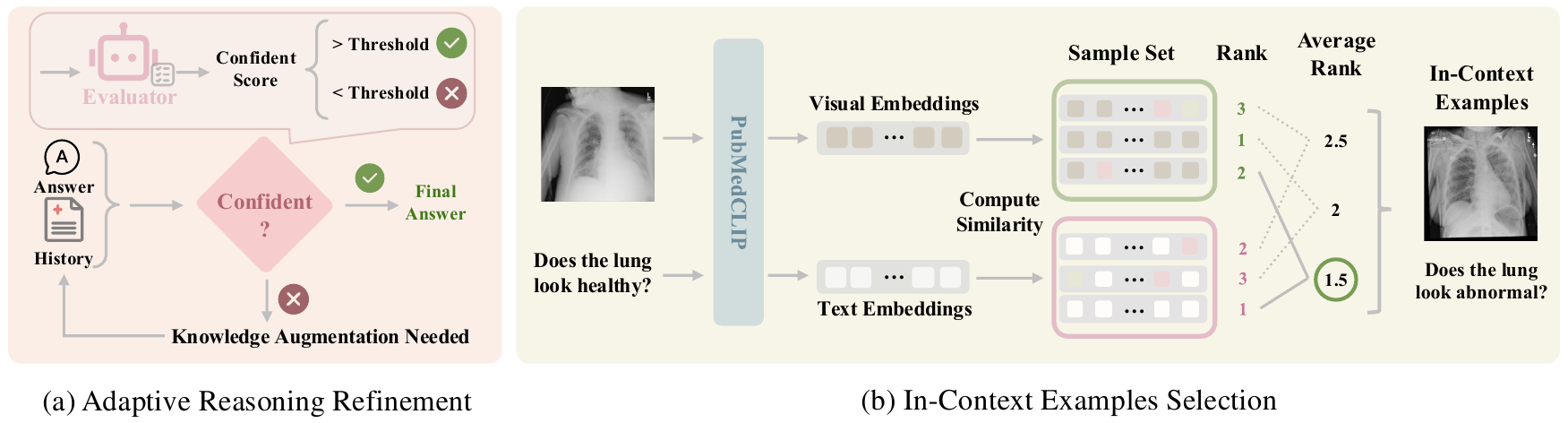}
\caption{\textbf{(a) Adaptive Reasoning Refinement:} The \texttt{Evaluator} agent dynamic controls the medical knowledge augmentation process by analyzing the consistency between the current answer and accumulated reasoning history. \textbf{(b) In-Context Examples Selection:} The system ranks candidate examples using a dual-similarity metric combining visual and textual features, selecting top-K examples as in-context examples.}
\label{fig: icl}  
\vspace{-0.3cm}
\end{figure*}

This coarse-to-fine approach enhances the intrinsic medical reasoning capability of Med-MLLMs in two ways: (1) breaking down complex analyses into focused steps through hierarchical questioning, enabling thorough examination of diagnostic features; and (2) building a clear reasoning chain that progressively refines visual understanding. This hierarchical questioning strategy is designed to mimic the systematic diagnostic process employed by clinicians, progressing from general impressions to detailed anatomical inspection and finally to the identification of specific pathologies. This clinical workflow alignment makes our approach particularly effective for medical tasks, distinguishing it from generic question decomposition methods. Through this progressive analysis, we effectively guide Med-MLLMs to uncover their intrinsic medical knowledge and generate more accurate and detailed diagnostic insights.


\noindent\textbf{Extrinsic Medical Knowledge Augmentation.}
While our intrinsic Med-KA enhances the depth of medical visual reasoning, Med-MLLMs still face the extrinsic medical reasoning bottleneck due to their static pre-trained knowledge. This issue is particularly critical in data-efficient scenarios where models encounter novel cases that require specialized medical expertise. Without comprehensive domain knowledge, models often generate plausible but incorrect responses, leading to potential hallucinations~\cite{xia2024rule, yan2024worse}.

To address this remaining challenge, we introduce an extrinsic Med-KA strategy accomplished by our \texttt{Retriever} agent. Inspired by recent advances in Retrieval Augmented Generation~\cite{soman2024biomedical, xiong2024benchmarking}, our approach consists of two steps. First, the \texttt{Retriever} agent uses an LLM to analyze the accumulated context (including medical captions, questions, and reasoning history) to extract key medical concepts such as "pulmonary nodule". These concepts then serve as queries to SPOKE~\cite{morris2023scalable}, a comprehensive biomedical knowledge graph containing 42 million nodes and 160 million edges assembled from 41 different biomedical databases. Through SPOKE queries, the \texttt{Retriever} agent obtains relevant subgraphs containing structured medical knowledge, including disease-symptom associations, anatomical relationships, and medical presentations. These medical facts are then transformed into natural language descriptions for integration into the reasoning process to ground the medical diagnosis. To ensure the quality of retrieved knowledge and mitigate the risk of introducing noise, our \texttt{Retriever} employs a multi-faceted filtering approach. It first extracts precise medical concepts to query the knowledge graph, and then filters the retrieved facts based on embedding similarity with the query context, ensuring only the most relevant information is integrated into the reasoning process.

This extrinsic Med-KA mechanism strengthens Med-MLLMs' reasoning reliability in two ways. First, by retrieving relevant medical knowledge from an external medical knowledge graph, we provide models with specialized expertise needed for novel cases in data-efficient scenarios. Second, the retrieved structured medical facts serve as reliable domain expertise to ground the reasoning process, effectively reducing hallucinations. Together with intrinsic Med-KA, this approach enables Med-MLLMs to perform more reliable medical reasoning through both deeper visual analysis and grounded domain knowledge, especially in data-efficient scenarios.


\subsection{Implementation Extensions} \label{sec: extension}
Building upon our Med-KA mechanisms, we introduce two extensions to further enhance our framework's effectiveness and efficiency: an adaptive reasoning refinement mechanism, and a few-shot enhancement strategy.

\noindent\textbf{Adaptive Reasoning Refinement.} While our two Med-KA mechanisms enhance medical reasoning capabilities, they often require multiple iterations of analysis to achieve comprehensive understanding. However, we observe that excessive refinement can be counterproductive (shown in Fig.~\ref{fig: experiments_comparison}(a): continuous accumulation of information beyond what's necessary may introduce noise and inconsistencies, potentially overturning initially correct judgments. Moreover, unnecessary iterations increase computational overhead without proportional gains in accuracy. To balance reasoning thoroughness with computational efficiency, we introduce an adaptive reasoning refinement mechanism, implemented through our \texttt{Evaluator} agent (Fig.\ref{fig: icl}(a)). The \texttt{Evaluator} dynamically controls the knowledge augmentation process by analyzing the consistency between current answers and accumulated reasoning history. It computes a confidence score based on predefined criteria (detailed in Appendix\ref{appendix: prompt}). When this score exceeds a threshold of 3 out of 5—indicating sufficient reasoning depth and reliability—the system concludes its analysis. If the maximum iteration limit is reached without meeting the confidence threshold, the system adopts the final iteration's response. This adaptive control prevents excessive refinement while ensuring accurate and efficient medical reasoning.

\noindent\textbf{Few-Shot Enhancement.} To further demonstrate our framework's effectiveness in data-efficient settings, we extend it to few-shot scenarios via in-context learning. The key challenge lies in selecting the most relevant examples that can effectively guide the reasoning process. To address this, we propose a \textit{dual-similarity selection strategy}. As illustrated in Fig.~\ref{fig: icl}(b), we utilize PubMedCLIP~\citep{zhang2023large} to compute similarities in both textual and visual domains. Formally, given a test sample with question embedding $\mathcal{T}$ and image embedding $\mathcal{I}$, we select the top $K$ examples from a candidate sample set $M$ through:
\[
\operatorname{ICL}_K = \operatorname{TopK}_{i \in M} \frac{1}{2} \left( \operatorname{sim}(\mathcal{T}, \mathcal{T}_i) + \operatorname{sim}(\mathcal{I}, \mathcal{I}_i) \right)
\]
where $\operatorname{ICL}_K = \{(c_k, q_k, \hat{a}_k)\}_{k=1}^K$ represents the selected examples containing caption, question, and answer triplets. The caption $c_k$ is generated by the \texttt{Perceiver} agent from the corresponding medical image. These carefully chosen examples are integrated into our framework, enabling the \textit{Reasoner} to leverage similar cases for more accurate diagnosis. This extension demonstrates our framework's adaptability across both zero-shot and few-shot settings, highlighting its effectiveness in data-efficient medical visual reasoning.

\begin{table*}[t!]
    \centering
    \tiny
    \renewcommand{\arraystretch}{1.2}
    \setlength{\tabcolsep}{1.8pt}
    \begin{tabular}{l c c| c c| c |c |c |c |c |c c}
        \toprule
        \multirow{2}{*}{\textbf{Method}} & \multicolumn{2}{c}{\textbf{VQA-RAD}} & \multicolumn{2}{c}{\textbf{SLAKE}} & \multicolumn{1}{c}{\textbf{IU-Xray}} & \multicolumn{1}{c}{\textbf{OL3I}} & \multicolumn{1}{c}{\textbf{OmniMedVQA}} & \multicolumn{1}{c}{\textbf{FairVL-Med}} & \multicolumn{1}{c}{\textbf{PMC-OA}} & \multirow{2}{*}{\textbf{Average}} \\
        \cmidrule{2-10}
        & \textbf{Open} & \textbf{Closed} & \textbf{Open} & \textbf{Closed} & \textbf{Closed} & \textbf{Closed} & \textbf{Closed} & \textbf{Open} & \textbf{Open} & \\
        \midrule
        
        \textbf{LLaVA-Med-v1.5} & 30.50 & 52.94 & 41.74 & 44.95 & 34.50 & 22.80 & 40.30 & 54.58 & 56.46 & 42.09 \\
        \hspace{4pt}\textbf{+ Img2LLM} & 37.81 {\tiny\textcolor[HTML]{AF6673}{(+7.31)}} & 47.43 {\tiny\textcolor[HTML]{7c9c5c}{(-5.51)}} & 50.89 {\tiny\textcolor[HTML]{AF6673}{(+9.15)}} & 59.86 {\tiny\textcolor[HTML]{AF6673}{(+14.91)}} & 70.60 {\tiny\textcolor[HTML]{AF6673}{(+36.10)}} & 49.80 {\tiny\textcolor[HTML]{AF6673}{(+27.00)}} & 54.40 {\tiny\textcolor[HTML]{AF6673}{(+14.10)}} & 61.74 {\tiny\textcolor[HTML]{AF6673}{(+7.16)}} & 63.03 {\tiny\textcolor[HTML]{AF6673}{(+6.57)}} & 55.06 {\tiny\textcolor[HTML]{AF6673}{(+12.97)}} \\
        \hspace{4pt}\textbf{+ IdealGPT} & 41.56 {\tiny\textcolor[HTML]{AF6673}{(+11.06)}} & 61.40 {\tiny\textcolor[HTML]{AF6673}{(+8.46)}} & 50.96 {\tiny\textcolor[HTML]{AF6673}{(+9.22)}} & 69.95 {\tiny\textcolor[HTML]{AF6673}{(+25.00)}} & 67.80 {\tiny\textcolor[HTML]{AF6673}{(+33.30)}} & 65.40 {\tiny\textcolor[HTML]{AF6673}{(+42.60)}} & 53.90 {\tiny\textcolor[HTML]{AF6673}{(+13.60)}} & 63.13 {\tiny\textcolor[HTML]{AF6673}{(+8.55)}} & 68.02 {\tiny\textcolor[HTML]{AF6673}{(+11.56)}} & 60.23 {\tiny\textcolor[HTML]{AF6673}{(+18.14)}} \\
        \rowcolor{mygray}\hspace{4pt}\textbf{+ \shortname{}} & \textbf{42.19} {\tiny\textcolor[HTML]{AF6673}{(+11.69)}} & 61.03 {\tiny\textcolor[HTML]{AF6673}{(+8.09)}} & 54.39 {\tiny\textcolor[HTML]{AF6673}{(+12.65)}} & 70.43 {\tiny\textcolor[HTML]{AF6673}{(+25.48)}} & 70.30 {\tiny\textcolor[HTML]{AF6673}{(+35.80)}} & 65.40 {\tiny\textcolor[HTML]{AF6673}{(+42.60)}} & 57.20 {\tiny\textcolor[HTML]{AF6673}{(+16.90)}} & 66.60 {\tiny\textcolor[HTML]{AF6673}{(+12.02)}} & 65.51 {\tiny\textcolor[HTML]{AF6673}{(+9.05)}} & 61.45 {\tiny\textcolor[HTML]{AF6673}{(+19.36)}} \\
        \rowcolor{mygray}\hspace{4pt}\textbf{+ \shortname{} w/ FS} & 41.73 {\tiny\textcolor[HTML]{AF6673}{(+11.23)}} & \textbf{63.97} {\tiny\textcolor[HTML]{AF6673}{(+11.03)}} & \textbf{54.41} {\tiny\textcolor[HTML]{AF6673}{(+12.67)}} & \textbf{73.56} {\tiny\textcolor[HTML]{AF6673}{(+28.61)}} & \textbf{70.80} {\tiny\textcolor[HTML]{AF6673}{(+36.30)}} & \textbf{67.00} {\tiny\textcolor[HTML]{AF6673}{(+44.20)}} & \textbf{62.20} {\tiny\textcolor[HTML]{AF6673}{(+21.90)}} & \textbf{66.85} {\tiny\textcolor[HTML]{AF6673}{(+12.27)}} & \textbf{65.76} {\tiny\textcolor[HTML]{AF6673}{(+9.30)}} & \textbf{62.92} {\tiny\textcolor[HTML]{AF6673}{(+20.83)}} \\
        \midrule
        \textbf{Med-InstructBLIP} & 32.41 & 61.76 & 42.82 & 59.38 & 68.60 & 34.40 & 29.50 & 52.18 & 57.85 & 48.77 \\
        \hspace{4pt}\textbf{+ Img2LLM} & 37.61 {\tiny\textcolor[HTML]{AF6673}{(+5.20)}} & 57.72 {\tiny\textcolor[HTML]{7c9c5c}{(-4.04)}} & 47.33 {\tiny\textcolor[HTML]{AF6673}{(+4.51)}} & 69.23 {\tiny\textcolor[HTML]{AF6673}{(+9.85)}} & 73.10 {\tiny\textcolor[HTML]{AF6673}{(+4.50)}} & 46.00 {\tiny\textcolor[HTML]{AF6673}{(+11.60)}} & 59.60 {\tiny\textcolor[HTML]{AF6673}{(+30.10)}} & 59.75 {\tiny\textcolor[HTML]{AF6673}{(+7.57)}} & 56.39 {\tiny\textcolor[HTML]{7c9c5c}{(-1.46)}} & 56.30 {\tiny\textcolor[HTML]{AF6673}{(+7.53)}} \\
        \hspace{4pt}\textbf{+ IdealGPT} & 40.22 {\tiny\textcolor[HTML]{AF6673}{(+7.81)}} & 65.07 {\tiny\textcolor[HTML]{AF6673}{(+3.31)}} & 48.85 {\tiny\textcolor[HTML]{AF6673}{(+6.03)}} & 65.14 {\tiny\textcolor[HTML]{AF6673}{(+5.76)}} & 80.70 {\tiny\textcolor[HTML]{AF6673}{(+12.10)}} & 67.40 {\tiny\textcolor[HTML]{AF6673}{(+33.00)}} & 56.30 {\tiny\textcolor[HTML]{AF6673}{(+26.80)}} & 64.12 {\tiny\textcolor[HTML]{AF6673}{(+11.94)}} & 60.10 {\tiny\textcolor[HTML]{AF6673}{(+2.25)}} & 60.88 {\tiny\textcolor[HTML]{AF6673}{(+12.11)}} \\
        \rowcolor{mygray}\hspace{4pt}\textbf{+ \shortname{}} & 41.02 {\tiny\textcolor[HTML]{AF6673}{(+8.61)}} & 68.75 {\tiny\textcolor[HTML]{AF6673}{(+6.99)}} & 51.13 {\tiny\textcolor[HTML]{AF6673}{(+8.31)}} & 69.47 {\tiny\textcolor[HTML]{AF6673}{(+10.09)}} & 79.50 {\tiny\textcolor[HTML]{AF6673}{(+10.90)}} & \textbf{67.60} {\tiny\textcolor[HTML]{AF6673}{(+33.20)}} & 62.70 {\tiny\textcolor[HTML]{AF6673}{(+33.20)}} & 66.61 {\tiny\textcolor[HTML]{AF6673}{(+14.43)}} & 63.97 {\tiny\textcolor[HTML]{AF6673}{(+6.12)}} & 63.42 {\tiny\textcolor[HTML]{AF6673}{(+14.65)}} \\
        \rowcolor{mygray}\hspace{4pt}\textbf{+ \shortname{} w/ FS} & \textbf{46.75} {\tiny\textcolor[HTML]{AF6673}{(+14.34)}} & \textbf{74.26} {\tiny\textcolor[HTML]{AF6673}{(+12.50)}} & \textbf{52.03} {\tiny\textcolor[HTML]{AF6673}{(+9.21)}} & \textbf{72.84} {\tiny\textcolor[HTML]{AF6673}{(+13.46)}} & \textbf{84.90} {\tiny\textcolor[HTML]{AF6673}{(+16.30)}} & 67.00 {\tiny\textcolor[HTML]{AF6673}{(+32.60)}} & \textbf{71.20} {\tiny\textcolor[HTML]{AF6673}{(+41.70)}} & \textbf{67.10} {\tiny\textcolor[HTML]{AF6673}{(+12.98)}} & \textbf{65.74} {\tiny\textcolor[HTML]{AF6673}{(+7.89)}} & \textbf{66.87} {\tiny\textcolor[HTML]{AF6673}{(+18.10)}} \\
        \midrule
        \textbf{Med-BLIVA} & 29.19 & 61.76 & 43.51 & 56.01 & 69.80 & 38.20 & 31.90 & 49.33 & 54.41 & 48.24 \\
        \hspace{4pt}\textbf{+ Img2LLM} & 32.76 {\tiny\textcolor[HTML]{AF6673}{(+3.57)}} & 59.93 {\tiny\textcolor[HTML]{7c9c5c}{(-1.83)}} & 44.95 {\tiny\textcolor[HTML]{AF6673}{(+1.44)}} & 62.74 {\tiny\textcolor[HTML]{AF6673}{(+6.73)}} & 70.10 {\tiny\textcolor[HTML]{AF6673}{(+0.30)}} & 46.20 {\tiny\textcolor[HTML]{AF6673}{(+8.00)}} & 57.80 {\tiny\textcolor[HTML]{AF6673}{(+25.90)}} & 62.43 {\tiny\textcolor[HTML]{AF6673}{(+13.10)}} & 55.69 {\tiny\textcolor[HTML]{AF6673}{(+1.28)}} & 55.27 {\tiny\textcolor[HTML]{AF6673}{(+7.03)}} \\
        \hspace{4pt}\textbf{+ IdealGPT} & 40.84 {\tiny\textcolor[HTML]{AF6673}{(+11.65)}} & 53.31 {\tiny\textcolor[HTML]{7c9c5c}{(-8.45)}} & 50.08 {\tiny\textcolor[HTML]{AF6673}{(+6.57)}} & 64.66 {\tiny\textcolor[HTML]{AF6673}{(+8.65)}} & 71.40 {\tiny\textcolor[HTML]{AF6673}{(+1.60)}} & 47.20 {\tiny\textcolor[HTML]{AF6673}{(+9.00)}} & 57.80 {\tiny\textcolor[HTML]{AF6673}{(+25.90)}} & 64.94 {\tiny\textcolor[HTML]{AF6673}{(+15.61)}} & 61.30 {\tiny\textcolor[HTML]{AF6673}{(+6.89)}} & 56.84 {\tiny\textcolor[HTML]{AF6673}{(+8.60)}} \\
        \rowcolor{mygray}\hspace{4pt}\textbf{+ \shortname{}} & 41.40 {\tiny\textcolor[HTML]{AF6673}{(+12.21)}} & 61.76  {\tiny\textcolor[HTML]{AF6673}{(+0.00)}} & \textbf{50.95} {\tiny\textcolor[HTML]{AF6673}{(+7.44)}} & 68.75 {\tiny\textcolor[HTML]{AF6673}{(+12.74)}} & 76.70 {\tiny\textcolor[HTML]{AF6673}{(+6.90)}} & \textbf{67.00} {\tiny\textcolor[HTML]{AF6673}{(+28.80)}} & 63.20 {\tiny\textcolor[HTML]{AF6673}{(+31.30)}} & 66.61 {\tiny\textcolor[HTML]{AF6673}{(+17.28)}} & 63.97 {\tiny\textcolor[HTML]{AF6673}{(+9.56)}} & 62.26 {\tiny\textcolor[HTML]{AF6673}{(+14.02)}} \\
        \rowcolor{mygray}\hspace{4pt}\textbf{+ \shortname{} w/ FS} & \textbf{45.16} {\tiny\textcolor[HTML]{AF6673}{(+15.97)}} & \textbf{67.65} {\tiny\textcolor[HTML]{AF6673}{(+5.89)}} & 50.49 {\tiny\textcolor[HTML]{AF6673}{(+6.98)}} & \textbf{69.23} {\tiny\textcolor[HTML]{AF6673}{(+13.22)}} & \textbf{84.60} {\tiny\textcolor[HTML]{AF6673}{(+14.80)}} & 65.80 {\tiny\textcolor[HTML]{AF6673}{(+27.60)}} & \textbf{65.90} {\tiny\textcolor[HTML]{AF6673}{(+34.00)}} & \textbf{67.10} {\tiny\textcolor[HTML]{AF6673}{(+17.77)}} & \textbf{65.74} {\tiny\textcolor[HTML]{AF6673}{(+11.33)}} & \textbf{64.63} {\tiny\textcolor[HTML]{AF6673}{(+16.39)}} \\
        \bottomrule
    \end{tabular}
    \caption{\textbf{Zero-shot and Few-shot Performance Comparison.} Our framework consistently improves the performance of different Med-MLLMs across various benchmarks. FS denotes experiments with 4 in-context examples.}
    \label{tab: med-mllm}
    \vspace{-0.3cm}
\end{table*}

%% file: sec/4_experiments.tex
\section{Experiments} \label{sec: exp}

\subsection{Experimental Details}
\noindent\textbf{Experimental Setup.} We evaluate \shortname{} on eight Med-VQA benchmarks that cover diverse medical domains and imaging modalities (detailed in Appendix~\ref{appendix: benchmarks}). For evaluation models, we primarily use LLaVA-Med-v1.5~\cite{li2024llava}. We also develop variants of Med-InstructBLIP~\cite{instructblip} and Med-BLIVA~\cite{hu2024bliva} both using LLaMA-v1 as their LLM backbone and following LLaVA-Med's training methodology (detailed in Appendix~\ref{appendix: models}). Following prior work~\cite{li2024llava}, we use accuracy for closed-ended questions and recall for open-ended questions. Additional experiments with general-purpose MLLMs are provided in Appendix~\ref{appendix: general-mllm}.

\vspace{-0.1cm}
\noindent\textbf{Baselines.} We compare \shortname{} with three types of approaches: (1) Single-step inference by Med-MLLMs serving as our zero-shot baseline; (2) Two-stage methods such as Img2LLM~\cite{guo2023images}, which generate image captions via MLLMs before LLM reasoning; and (3) Agent-based approaches like IdealGPT~\cite{you2023idealgpt} that utilize multiple LLMs for collaborative reasoning.

\vspace{-0.1cm}
\noindent\textbf{Implementation Details.} Our framework uses GPT-4o as the core reasoning engine for all agents by default. For adaptive reasoning refinement, we set a maximum of 3 iterations and a confidence threshold of 3/5. For few-shot experiments, we use 4 in-context examples as the default setting.

\subsection{Effectiveness of \shortname{}}

\noindent\textbf{Zero-shot Med-VQA.} As shown in Table~\ref{tab: med-mllm} demonstrates the substantial improvements achieved by our framework across different Med-MLLMs and evaluation benchmarks. With LLaVA-Med-v1.5~\cite{li2024llava}, \shortname{} achieves an average improvement of \textbf{19.36\%} over the direct inference baseline. Using Med-BLIVA~\cite{hu2024bliva}, our method outperforms existing LLM-empowered approaches like Img2LLM~\cite{guo2023images} and IdealGPT~\cite{you2023idealgpt} by \textbf{6.36\%} and \textbf{5.42\%} respectively. These significant improvements stem from our medical-specific design choices. While Img2LLM~\cite{guo2023images} only relies on caption generation and IdealGPT~\cite{you2023idealgpt} uses general-purpose agent collaboration, our framework enhances medical reasoning through both intrinsic and extrinsic Med-KA along with adaptive reasoning refinement.

\noindent\textbf{Few-shot Med-VQA.} We further enhance our framework's effectiveness through few-shot learning, enabling performance gains without model fine-tuning. As shown in Table~\ref{tab: med-mllm}, this few-shot enhancement leads to consistent improvements across all benchmarks, with Med-InstrcuctBLIP achieving a further \textbf{3.45\%} gain over its zero-shot performance. These improvements demonstrate the effectiveness of our dual-similarity selection strategy, which provides the \texttt{Reasoner} with highly relevant in-context examples to strengthen its medical reasoning capability. These results highlight \shortname{}'s strong adaptability in data-efficient scenarios, from zero-shot to few-shot settings.

\begin{table}[t]
    \centering
    \scriptsize
    \renewcommand{\arraystretch}{1.2}
    \setlength{\tabcolsep}{1pt}
    \begin{tabular}{l c c c| c }
        \toprule
        \multirow{2}{*}{\textbf{Model}} & \multicolumn{3}{c}{\textbf{Hallucination Question Type}} & \multirow{2}{*}{\textbf{Average}} \\
        \cmidrule{2-4}
        & \textbf{Organ} & \textbf{Condition} & \textbf{Abnormality} & \\
        \midrule
        
        \textbf{LLaVA-Med-v1.5} & 39.60 & 30.30 & 21.96 & 30.62 \\
        \rowcolor{mygray} \hspace{4pt}\textbf{+ \shortname{}} & 88.00 {\tiny\textcolor[HTML]{AF6673}{(+48.40)}} & 91.80 {\tiny\textcolor[HTML]{AF6673}{(+61.50)}} & 54.00 {\tiny\textcolor[HTML]{AF6673}{(+32.04)}} & 77.93 {\tiny\textcolor[HTML]{AF6673}{(+47.31)}} \\
        \rowcolor{mygray} \hspace{4pt}\textbf{+ \shortname{} w/ FS} & \textbf{92.40} {\tiny\textcolor[HTML]{AF6673}{(+52.80)}} & \textbf{94.80} {\tiny\textcolor[HTML]{AF6673}{(+64.50)}} & \textbf{54.40} {\tiny\textcolor[HTML]{AF6673}{(+32.44)}} & \textbf{80.53} {\tiny\textcolor[HTML]{AF6673}{(+49.91)}} \\
        \midrule
        
        \textbf{Med-InstructBLIP} & 37.20 & 16.60 & 60.60 & 38.13 \\
        \rowcolor{mygray} \hspace{4pt}\textbf{+ \shortname{}} & 89.80 {\tiny\textcolor[HTML]{AF6673}{(+52.60)}} & \textbf{94.00} {\tiny\textcolor[HTML]{AF6673}{(+77.40)}} & 64.40 {\tiny\textcolor[HTML]{AF6673}{(+3.80)}} & 82.73 {\tiny\textcolor[HTML]{AF6673}{(+44.60)}} \\
        \rowcolor{mygray} \hspace{4pt}\textbf{+ \shortname{} w/ FS} & \textbf{92.00} {\tiny\textcolor[HTML]{AF6673}{(+54.80)}} & 93.00 {\tiny\textcolor[HTML]{AF6673}{(+76.40)}} & \textbf{65.60} {\tiny\textcolor[HTML]{AF6673}{(+5.00)}} & \textbf{83.53} {\tiny\textcolor[HTML]{AF6673}{(+45.40)}} \\
        \midrule
        
        \textbf{Med-BLIVA} & 65.80 & 53.60 & 61.80 & 60.40 \\
        \rowcolor{mygray} \hspace{4pt}\textbf{+ \shortname{}} & 83.80 {\tiny\textcolor[HTML]{AF6673}{(+18.00)}} & 87.80 {\tiny\textcolor[HTML]{AF6673}{(+34.20)}} & 61.20 {\tiny\textcolor[HTML]{AF6673}{(-0.60)}} & 77.60 {\tiny\textcolor[HTML]{AF6673}{(+17.20)}} \\
        \rowcolor{mygray} \hspace{4pt}\textbf{+ \shortname{} w/ FS} & \textbf{90.60} {\tiny\textcolor[HTML]{AF6673}{(+24.80)}} & \textbf{92.80} {\tiny\textcolor[HTML]{AF6673}{(+39.20)}} & \textbf{64.20} {\tiny\textcolor[HTML]{AF6673}{(+2.40)}} & \textbf{82.53} {\tiny\textcolor[HTML]{AF6673}{(+22.13)}} \\
        \bottomrule
    \end{tabular}
    \caption{\textbf{Effectiveness in reducing hallucination.}}
    \label{tab:hallucination}
    \vspace{-0.5cm}
\end{table}

\noindent\textbf{Medical Hallucination Reduction.} Beyond improving overall performance, a critical measure of our framework's effectiveness lies in reducing medical hallucinations. We evaluate this capability using ProbMed~\cite{yan2024worse}, a specialized benchmark for assessing models' medical reasoning reliability. As shown in Table~\ref{tab:hallucination}, \shortname{} achieves substantial reductions in hallucination rates across all tested models, with Med-InstructBLIP~\cite{instructblip} achieving a \textbf{47.37\%} reduction. These results demonstrate that our intrinsic and extrinsic Med-KA effectively grounds the medical reasoning process with reliable domain knowledge, addressing a crucial challenge in real-world clinical applications.

\subsection{Further Analysis}

\begin{figure}[t!]
\vspace{-0.16cm}
\centering  
\includegraphics[height=13cm]{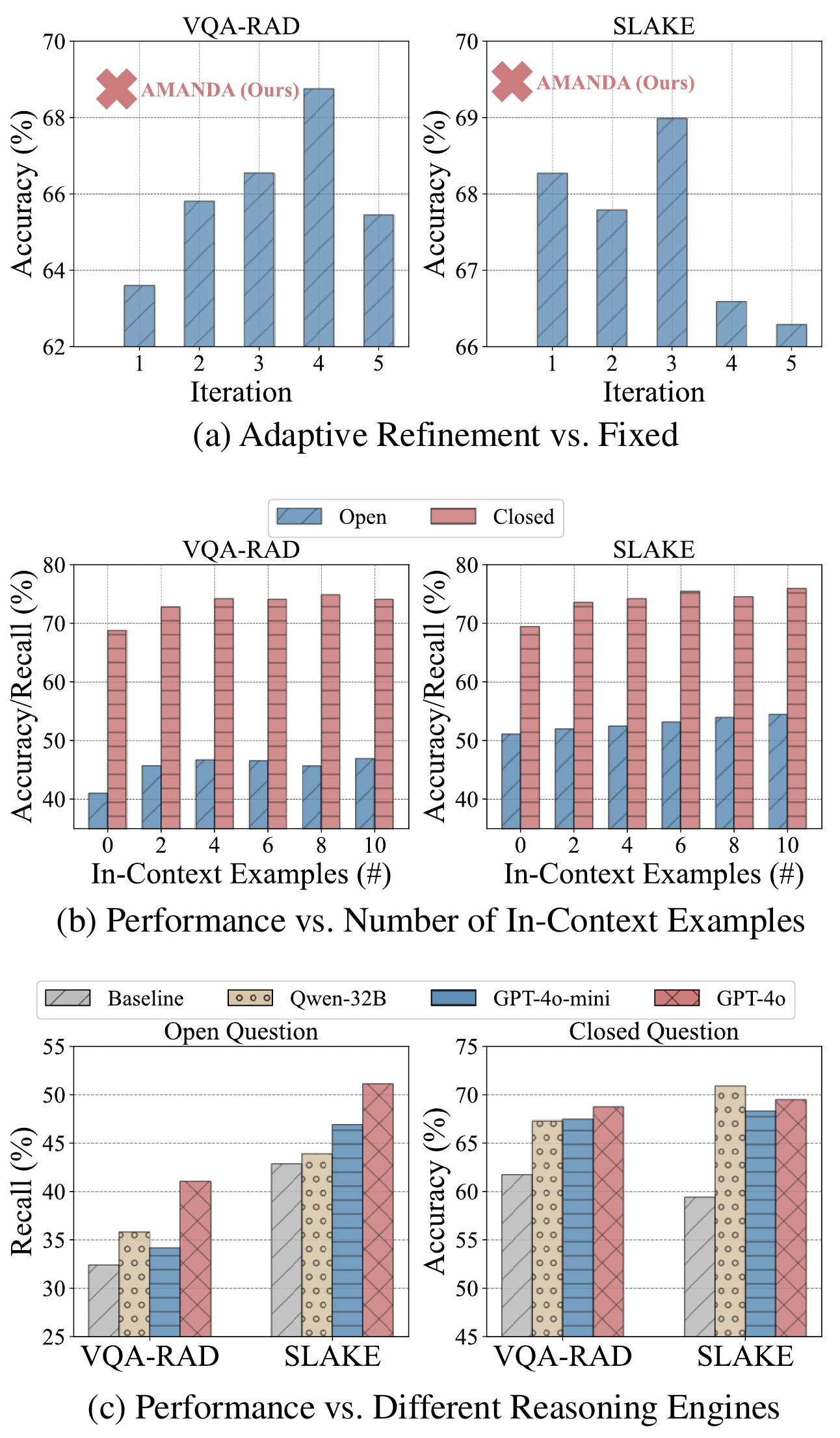}
\caption{Analysis of framework components.}
\label{fig: experiments_comparison}  
\vspace{-0.3cm}
\end{figure}

\noindent\textbf{Effectiveness of Adaptive Refinement.} Fig.~\ref{fig: experiments_comparison}(a) demonstrates the superiority of our adaptive approach over fixed-iteration strategies. In fixed-iteration settings, performance initially improves with additional iterations but eventually degrades, revealing the detrimental effects of excessive refinement. Our adaptive mechanism achieves dual benefits: it increases accuracy from 66.54\% to 68.75\% while reducing the average number of iterations from 3.0 to 0.61, resulting in approximately \textbf{4.9x} improved efficiency.

\noindent\textbf{Number of In-Context Examples.} Fig.~\ref{fig: experiments_comparison}(b) illustrates how the number of in-context examples affects model performance. While increasing examples initially improves results, the benefits plateau beyond an optimal point. This finding suggests that carefully selected examples are more crucial than quantity for enhancing medical reasoning.

\noindent\textbf{Reasoning Engines Compatibility.} As shown in Fig.~\ref{fig: experiments_comparison}(c), our framework demonstrates compatibility with both closed-source (GPT-4o, GPT-4o-mini) and open-source (DeepSeek-R1-Distill-Qwen-32B~\cite{guo2025deepseek}) LLMs as reasoning engines. GPT-4o achieves superior performance on open-ended questions, while open-source alternatives like DeepSeek-R1-Distill-Qwen-32B show competitive results on closed-ended questions. This versatility highlights our method's adaptability across different reasoning engines, enabling users to balance performance requirements with computational cost considerations.

\begin{table}[t]
    \centering
    \scriptsize
    \renewcommand{\arraystretch}{1.2}
    \setlength{\tabcolsep}{2.2pt}
    \begin{tabular}{c c c | c c | c c }
        \toprule
        \multirow{2}{*}{\textbf{Model}} & \multirow{2}{*}{\textbf{Model Size}} & \multirow{2}{*}{\textbf{Dataset Size}} & \multicolumn{2}{c}{\textbf{VQA-RAD}} & \multicolumn{2}{c}{\textbf{SLAKE}} \\
        \cmidrule{4-7}
        & & & \textbf{Open} & \textbf{Closed} & \textbf{Open} & \textbf{Closed} \\
        \midrule
        \textbf{LLaMA} & 7B & 60K & 41.40 & 61.76 & 50.95 & 68.75 \\
        \textbf{LLaMA} & 13B & 60K & 38.34 & \textbf{66.54} & \underline{51.85} & \textbf{69.47} \\
        \textbf{LLaMA} & 7B & 150K & \textbf{47.90} & \underline{66.18} & 51.25 & 68.27 \\
        \textbf{Vicuna} & 7B & 60K & \underline{41.63} & 58.82 & \textbf{51.90} & 67.31 \\
        \textbf{PMC-LLaMA} & 7B & 60K & 40.80 & 62.87 & 51.01 & \underline{68.75} \\
        \bottomrule
    \end{tabular}
        \vspace{-0.2cm}
    \caption{\textbf{Analysis of language backbones in Med-BLIVA.} Each column’s highest score is in \textbf{bold}, while the second highest score is \underline{underlined}.}
    \label{tab:llmbackbone}
    \vspace{-0.1cm}
\end{table}

\begin{table}[t!]
    \centering
    \scriptsize
    \renewcommand{\arraystretch}{1.2}
    \setlength{\tabcolsep}{3pt}
    \begin{tabular}{l c c| c c}
        \toprule
        \multirow{2}{*}{\textbf{Method}} & \multicolumn{2}{c}{\textbf{VQA-RAD}} & \multicolumn{2}{c}{\textbf{SLAKE}} \\
        \cmidrule{2-5}
        & \textbf{Open} & \textbf{Closed} & \textbf{Open} & \textbf{Closed} \\
        \midrule
        \textbf{\shortname{}} & 42.19 & 61.03 & 54.39 & 70.43 \\
        \hspace{2pt}\textbf{- Perceiver} & 22.70 {\tiny\textcolor[HTML]{7c9c5c}{(-19.49)}} & 40.81 {\tiny\textcolor[HTML]{7c9c5c}{(-20.22)}} & 28.72 {\tiny\textcolor[HTML]{7c9c5c}{(-25.67)}} & 35.58 {\tiny\textcolor[HTML]{7c9c5c}{(-34.85)}} \\
        \hspace{2pt}\textbf{- Explorer} & 38.82 {\tiny\textcolor[HTML]{7c9c5c}{(-3.37)}} & 56.62 {\tiny\textcolor[HTML]{7c9c5c}{(-4.41)}} & 50.28 {\tiny\textcolor[HTML]{7c9c5c}{(-4.11)}} & 64.66 {\tiny\textcolor[HTML]{7c9c5c}{(-5.77)}} \\
        \hspace{2pt}\textbf{- Retriever} & 41.11 {\tiny\textcolor[HTML]{7c9c5c}{(-1.08)}} & 60.29 {\tiny\textcolor[HTML]{7c9c5c}{(-0.74)}} & 52.90 {\tiny\textcolor[HTML]{7c9c5c}{(-1.49)}} & 69.47 {\tiny\textcolor[HTML]{7c9c5c}{(-0.96)}} \\
        \hspace{2pt}\textbf{- Reasoner} & 38.09 {\tiny\textcolor[HTML]{7c9c5c}{(-4.10)}} & 57.72 {\tiny\textcolor[HTML]{7c9c5c}{(-3.31)}} & 50.21 {\tiny\textcolor[HTML]{7c9c5c}{(-4.18)}} & 68.03 {\tiny\textcolor[HTML]{7c9c5c}{(-2.40)}} \\
        \hspace{2pt}\textbf{- Evaluator} & 43.56 {\tiny\textcolor[HTML]{AF6673}{(+1.37)}} & 57.35 {\tiny\textcolor[HTML]{7c9c5c}{(-3.68)}} & 54.72 {\tiny\textcolor[HTML]{AF6673}{(+0.33)}} & 69.23 {\tiny\textcolor[HTML]{7c9c5c}{(-1.20)}} \\
        \bottomrule
    \end{tabular}
    \vspace{-0.2cm}
    \caption{\textbf{Ablation study.} Analysis of different agents by removing each from the full model.}
    \label{tab:ablation-study}
    \vspace{-0.4cm}
\end{table}

\noindent\textbf{Impact of MLLM Backbones.} Table~\ref{tab:llmbackbone} presents a comprehensive analysis of MLLMs with varying backbones and training configurations. Our evaluation reveals three key findings: \solidcircle{1} larger language backbones generally achieve better performance, particularly on closed-ended questions where precise reasoning is crucial; \solidcircle{2} increasing the pre-training dataset size from 60K~\cite{li2024llava} to 150K~\cite{cui2024biomedical} samples leads to significant improvements across all metrics; and \solidcircle{3} models with medical domain pre-training like PMC-LLaMA~\cite{wu2023pmc} demonstrate strong performance, highlighting the value of domain-specific knowledge in medical reasoning.

\subsection{Ablation Study}
To quantify the contribution of each agent, we conduct a systematic ablation study, with results presented in Tab.~\ref{tab:ablation-study}. The findings validate the integral role of each component in our framework. \noindent\textbf{Visual Components are Critical.} Removing the \texttt{Perceiver} or the \texttt{Explorer} leads to the most significant performance degradation. As shown in Tab.~\ref{tab:ablation-study}, removing the \texttt{Perceiver} causes a catastrophic performance drop of up to \textbf{34.85\%}, as it eliminates the foundational visual understanding. The \texttt{Explorer} is also crucial; its absence results in a drop of up to \textbf{5.77\%}, highlighting the importance of its coarse-to-fine decomposition for in-depth visual analysis. \noindent\textbf{Knowledge and Reasoning Components.} The \texttt{Retriever} and \texttt{Reasoner} also provide vital contributions. Disabling the \texttt{Retriever} is particularly detrimental for open-ended questions that require external knowledge, while removing the \texttt{Reasoner} impairs the framework's ability to synthesize information, leading to a performance drop of up to \textbf{4.18\%}. \noindent\textbf{The Role of the Evaluator.} The \texttt{Evaluator}'s role is nuanced. Removing it (i.e., running a fixed number of iterations) can slightly improve performance on some open-ended questions (e.g., \textbf{+1.37\%} on VQA-RAD) but degrades it on closed-ended ones, which are more susceptible to noise from excessive reasoning. However, its primary value lies in achieving the computational efficiency detailed in Sec.~4.3. These results collectively confirm that our dual approach—enhancing both intrinsic visual analysis and extrinsic knowledge grounding—is essential for achieving robust and accurate medical reasoning.

%% file: sec/5_conclusion.tex
\section{Conclusion}
In this work, we present \shortname{}, a training-free agentic framework that addresses Med-MLLMs' intrinsic and extrinsic bottlenecks in data-efficient scenarios. Our framework enhances medical visual reasoning through coarse-to-fine question decomposition and grounds its analysis with extrinsic knowledge graphs, while maintaining efficiency through adaptive reasoning refinement. Extensive experiments demonstrate substantial improvements on Med-VQA in both zero-shot and few-shot settings, highlighting \shortname{}'s potential for reliable AI-assisted medical diagnosis in resource-constrained environments.

%% file: sec/6_limitations.tex
\section{Limitations}
While our work demonstrates promising results, several perspectives remain for future exploration. First, although we evaluate on eight diverse Med-VQA benchmarks, testing on more specialized medical datasets across different modalities (e.g., MRI, CT) could further validate our framework's generalizability. Second, our experiments primarily focus on publicly available Med-MLLMs with language models up to 13B parameters; investigating the impact of larger language models (e.g., 70B) could potentially reveal additional performance gains. Third, incorporating more diverse external medical knowledge resources (e.g., medical textbooks, clinical guidelines, and medical reports) could potentially enhance our framework's capability in handling various types of medical queries. Fourth, enabling our agents to utilize existing medical tools and collaborate with hospitals for diagnosis would be a promising direction for real-world deployment. Finally, while we focus on a training-free approach, exploring lightweight fine-tuning strategies could potentially achieve better performance improvements while maintaining reasonable computational requirements in resource-constrained scenarios.

%% file: sec/X_suppl.tex
\section{Details of Evaluated MLLMs} \label{appendix: models}
We evaluate our framework across both medical domain-specific and general-domain MLLMs to demonstrate its versatility and effectiveness.
\subsection{Medical Domain-Specific MLLMs}
\begin{itemize}[leftmargin=*]
    \item \textbf{LLaVA-Med-v1.5}\cite{li2024llava}: Built on Mistral-7B\cite{jiang2023mistral}, this is our primary evaluation model. It extends LLaVA~\cite{liu2024visual} for medical domain understanding through specialized training on medical image-text pairs and conversational data.
    
    \item \textbf{Med-InstructBLIP}: Our medical adaptation of InstructBLIP~\cite{instructblip} using LLaMa-7B~\cite{touvron2023llama}. Following LLaVA-Med's training methodology~\cite{li2024llava}, we adapt the model for medical visual understanding while maintaining its instruction-tuning capabilities.
    
    \item \textbf{Med-BLIVA}: A medical version of BLIVA~\cite{hu2024bliva} based on LLaMa-7B~\cite{touvron2023llama}. We adapt it using LLaVA-Med's training strategy~\cite{li2024llava} to combine BLIVA's visual reasoning capabilities with medical domain expertise.
\end{itemize}

\subsection{Pre-training Details of Med-MLLMs}
For Med-InstructBLIP and Med-BLIVA, we follow LLaVA-Med's~\cite{li2024llava} two-stage training strategy:
\begin{itemize}[leftmargin=*]
\item \noindent\textbf{Stage 1: Feature Alignment.}
We first align the visual features with medical concepts through projection learning. Using 600K filtered image-text pairs from PMC-15M, we train only the projection layer while keeping both the visual encoder and language model frozen. This stage enables the models to understand biomedical visual concepts efficiently.

\item \noindent\textbf{Stage 2: Instruction Tuning.}
We then perform end-to-end instruction tuning with the projection layer and language model unfrozen. Using 60K medical image-text instruction data, we train the models to follow various medical instructions and perform visual reasoning tasks. This stage enhances the models' capabilities in medical visual understanding and dialogue interaction.
\end{itemize}

\subsection{General-Domain MLLMs}
\begin{itemize}[leftmargin=*]
    \item \textbf{InstructBLIP}~\cite{instructblip}: A strong general-domain MLLM with instruction-tuning capabilities. We evaluate it using its original pre-trained weights to assess our framework's effectiveness on models without medical domain adaptation.

    \item \textbf{xGen-MM}~\cite{xue2024xgen}: The latest BLIP architecture variant with advanced visual reasoning capabilities. We use its original weights to test our framework's compatibility with state-of-the-art general-purpose MLLMs.
\end{itemize}

Evaluating these general-domain models alongside medical-specific ones demonstrates our framework's versatility across different architectures and its ability to enhance medical reasoning capabilities regardless of domain specialization.

\section{Details of Med-VQA Benchmarks} \label{appendix: benchmarks}
\label{sec:dataset}
We utilize open-source Med-VQA benchmarks, which cover a wide range of medical image modalities and anatomical regions: VQA-RAD~\cite{lau2018dataset}, SLAKE~\cite{liu2021slake}, IU-Xray~\cite{demner2016preparing}, Harvard-FairVLMed~\cite{luo2024fairclip}, PMC-OA~\cite{lin2023pmc}, OL3I~\cite{zambrano2023opportunistic}, OmniMedVQA~\cite{hu2024omnimedvqa}, and ProbMed~\cite{yan2024worse}. Table~\ref{tab:dataset} provides comprehensive statistics about these datasets. The details of each benchmark are as follows:

\begin{table*}[t]
    \centering
    \caption{\textbf{Comprehensive statistics of the Med-VQA Benchmarks.}}
    \renewcommand{\arraystretch}{1.2}
    \resizebox{\linewidth}{!}{
    \begin{tabular}{c|l|cc|c|c|c|c}
        \toprule
        Index & Data Source & Modality & Region & \# Images & \# QA Items & Answer Type & \# Test \\\midrule
        1 & VQA-RAD~\cite{lau2018dataset} & X-Ray, CT & Chest, Abd & 315 & 3,515 & Mixed & 451 \\
        2 & SLAKE~\cite{liu2021slake} & CT, MRI, X-Ray & Mixture & 8,851 & 14,028 & Open-ended & 1,061 \\
        3 & IU-Xray~\cite{demner2016preparing} & X-Ray & Chest & 589 & 2,573 & Yes/No & 1,000 \\
        4 & Harvard-FairVLMed~\cite{luo2024fairclip} & Fundus & Eye & 713 & 2,838 & Open-ended & 1,000 \\
        5 & OL3I~\cite{zambrano2023opportunistic} & CT & Heart & 1,000 & 1,000 & Yes/No & 500 \\
        6 & PMC-OA~\cite{zhang2023pmc} & Mixture & Mixture & 2,587 & 13,294 & Open-ended & 1,000 \\
        7 & OmniMedVQA~\cite{hu2024omnimedvqa} & Mixture* & Mixture & 10,995 & 12,227 & Multi-choice & 1,000 \\
        8 & ProbMed~\cite{yan2024worse} & Mixture* & Mixture & 6,303 & 57,132 & Yes/No & 1,500 \\
    \bottomrule
    \end{tabular}
    }
    \label{tab:dataset}
\end{table*}

\begin{itemize}[leftmargin=*]
    \item \textbf{VQA-RAD}~\cite{lau2018dataset}: A dedicated Med-VQA dataset containing 315 medical images and 3,515 question-answer pairs. It covers various medical imaging modalities including chest X-rays and CT scans. The questions are carefully designed to evaluate both visual understanding and clinical reasoning capabilities, categorized into different types including modality, plane, organ system, and abnormality detection.

     \item \textbf{SLAKE}~\cite{liu2021slake}: A comprehensive Med-VQA dataset comprising 14,028 question-answer pairs on 8,851 medical images across multiple modalities (CT, MRI, X-Ray). The questions assess different levels of understanding, from basic pattern recognition to complex clinical reasoning. The dataset contains 11,222 training samples and 1,061 testing samples.
    
    \item \textbf{IU-Xray}~\cite{demner2016preparing}: A specialized dataset focusing on chest X-ray images and their corresponding diagnostic reports. Our benchmark includes 589 frontal chest X-rays from the test set, along with their detailed clinical reports.
    
    \item \textbf{Harvard-FairVLMed}~\cite{luo2024fairclip}: A multimodal dataset of fundus images designed to evaluate fairness in AI models. It contains image and text data from diverse demographic groups, specifically focusing on bias assessment in medical visual understanding.
    
    \item \textbf{PMC-OA}~\cite{lin2023pmc}: A large-scale collection of biomedical images extracted from open-access publications. We incorporate 2,587 diverse image-text pairs randomly selected from the test set into our benchmark.
    
    \item \textbf{OL3I}~\cite{zambrano2023opportunistic}: A publicly available dataset focused on predicting ischemic heart disease (IHD) using contrast-enhanced abdominal-pelvic CT examinations. It features a retrospective cohort with up to 5 years of follow-up data.
    
    \item \textbf{OmniMedVQA}~\cite{hu2024omnimedvqa}: A comprehensive Med-VQA benchmark collected from 73 different medical datasets. It encompasses 12 different imaging modalities and covers more than 20 distinct anatomical areas, providing broad coverage of medical visual understanding tasks.
    
    \item \textbf{ProbMed}~\cite{yan2024worse}: A specialized benchmark designed for evaluating model hallucination, comprising 6,303 images and 57,132 question-answer pairs. It includes carefully designed adversarial QA pairs across three modalities (X-ray, MRI, CT scan) and four anatomical regions (abdomen, brain, chest, spine).
\end{itemize}

\subsection{Evaluation Protocol}
Following \cite{xia2024cares}, we construct our evaluation benchmark using diverse medical image-text pairs from eight datasets. For classic Med-VQA benchmarks VQA-RAD and SLAKE, we use their complete test sets (451 and 1,061 QA pairs respectively) to maintain consistency with previous works. For larger-scale datasets (IU-Xray, Harvard-FairVLMed, OL3I, PMC-OA, OmniMedVQA, and ProbMed), we randomly sample 500-1,500 test examples from their original test sets due to computational constraints.

The remaining training samples from these datasets serve as our in-context learning pool for few-shot evaluation. For each test image, we retrieve similar examples based on visual and semantic similarity to construct few-shot prompts. This diverse collection of datasets, covering various modalities and answer formats (Yes/No, Open-ended, and Multi-choice), enables comprehensive evaluation of medical visual understanding capabilities.

\begin{table*}[t!]
    \centering
    \tiny
    \renewcommand{\arraystretch}{1.2}
    \setlength{\tabcolsep}{1.8pt}
    \begin{tabular}{l c c| c c| c |c |c |c |c |c c}
        \toprule
        \multirow{2}{*}{\textbf{Method}} & \multicolumn{2}{c}{\textbf{VQA-RAD}} & \multicolumn{2}{c}{\textbf{SLAKE}} & \multicolumn{1}{c}{\textbf{IU-Xray}} & \multicolumn{1}{c}{\textbf{OL3I}} & \multicolumn{1}{c}{\textbf{OmniMedVQA}} & \multicolumn{1}{c}{\textbf{FairVL-Med}} & \multicolumn{1}{c}{\textbf{PMC-OA}} & \multirow{2}{*}{\textbf{Average}} \\
        \cmidrule{2-10}
        & \textbf{Open} & \textbf{Closed} & \textbf{Open} & \textbf{Closed} & \textbf{Closed} & \textbf{Closed} & \textbf{Closed} & \textbf{Open} & \textbf{Open} & \\
        \midrule
\multicolumn{11}{c}{\it \textbf{General MLLMs (without Medical Pre-training)}} \\ \hline
\textbf{InstructBLIP} & 16.09 & 62.50 & 22.14 & 59.86 & 62.30 & 36.11 & 33.40 & 45.22 & 42.90 & 42.28 \\
\rowcolor{mygray}\hspace{4pt}\textbf{+ \shortname{}} & 29.86 {\tiny\textcolor[HTML]{AF6673}{(+13.77)}} & 65.81 {\tiny\textcolor[HTML]{AF6673}{(+3.31)}} & 41.03 {\tiny\textcolor[HTML]{AF6673}{(+18.89)}} & 66.35 {\tiny\textcolor[HTML]{AF6673}{(+6.49)}} & 68.30 {\tiny\textcolor[HTML]{AF6673}{(+6.00)}} & 61.11 {\tiny\textcolor[HTML]{AF6673}{(+25.00)}} & 52.30 {\tiny\textcolor[HTML]{AF6673}{(+18.90)}} & \textbf{64.83} {\tiny\textcolor[HTML]{AF6673}{(+19.61)}} & 63.08 {\tiny\textcolor[HTML]{AF6673}{(+20.18)}} & 56.96 {\tiny\textcolor[HTML]{AF6673}{(+14.68)}} \\
\rowcolor{mygray}\hspace{4pt}\textbf{+ \shortname{} w/ FS} & \textbf{38.96} {\tiny\textcolor[HTML]{AF6673}{(+22.87)}} & \textbf{68.01} {\tiny\textcolor[HTML]{AF6673}{(+5.51)}} & \textbf{48.61} {\tiny\textcolor[HTML]{AF6673}{(+26.47)}} & \textbf{69.71} {\tiny\textcolor[HTML]{AF6673}{(+9.85)}} & \textbf{71.30} {\tiny\textcolor[HTML]{AF6673}{(+9.00)}} & \textbf{63.89} {\tiny\textcolor[HTML]{AF6673}{(+27.78)}} & \textbf{54.40} {\tiny\textcolor[HTML]{AF6673}{(+21.00)}} & 64.81 {\tiny\textcolor[HTML]{AF6673}{(+19.59)}} & \textbf{63.12} {\tiny\textcolor[HTML]{AF6673}{(+20.22)}} & \textbf{60.31} {\tiny\textcolor[HTML]{AF6673}{(+18.03)}} \\
\midrule
\textbf{Xgen-MM} & 16.08 & 62.50 & 22.14 & 59.86 & 53.30 & 37.80 & 44.70 & 58.38 & 49.19 & 44.88 \\
\rowcolor{mygray}\hspace{4pt}\textbf{+ \shortname{}} & 35.20 {\tiny\textcolor[HTML]{AF6673}{(+19.12)}} & 67.28 {\tiny\textcolor[HTML]{AF6673}{(+4.78)}} & 46.47 {\tiny\textcolor[HTML]{AF6673}{(+24.33)}} & 70.19 {\tiny\textcolor[HTML]{AF6673}{(+10.33)}} & 59.20 {\tiny\textcolor[HTML]{AF6673}{(+5.90)}} & 48.80 {\tiny\textcolor[HTML]{AF6673}{(+11.00)}} & 54.10 {\tiny\textcolor[HTML]{AF6673}{(+9.40)}} & 67.34 {\tiny\textcolor[HTML]{AF6673}{(+8.96)}} & \textbf{64.85} {\tiny\textcolor[HTML]{AF6673}{(+15.66)}} & 57.05 {\tiny\textcolor[HTML]{AF6673}{(+12.17)}} \\
\rowcolor{mygray}\hspace{4pt}\textbf{+ \shortname{} w/ FS} & \textbf{37.76} {\tiny\textcolor[HTML]{AF6673}{(+21.68)}} & \textbf{75.37} {\tiny\textcolor[HTML]{AF6673}{(+12.87)}} & \textbf{47.92} {\tiny\textcolor[HTML]{AF6673}{(+25.78)}} & \textbf{74.28} {\tiny\textcolor[HTML]{AF6673}{(+14.42)}} & \textbf{69.60} {\tiny\textcolor[HTML]{AF6673}{(+16.30)}} & \textbf{51.60} {\tiny\textcolor[HTML]{AF6673}{(+13.80)}} & \textbf{58.10} {\tiny\textcolor[HTML]{AF6673}{(+13.40)}} & \textbf{67.42} {\tiny\textcolor[HTML]{AF6673}{(+9.04)}} & 64.72 {\tiny\textcolor[HTML]{AF6673}{(+15.53)}} & \textbf{60.75} {\tiny\textcolor[HTML]{AF6673}{(+15.87)}} \\
        \bottomrule
    \end{tabular}
\caption{\textbf{Generalization to general-purpose MLLMs.} Zero-shot and few-shot results across Med-VQA benchmarks using general MLLMs, showing the framework's strong generalization capability beyond Med-MLLMs.}
    \label{tab: general-mllm}
\end{table*}

\begin{table*}[t!]
    \centering
    \tiny
    \renewcommand{\arraystretch}{1.2}
    \setlength{\tabcolsep}{5pt}
    \begin{tabular}{l l c c| c c}
        \toprule
        \multirow{2}{*}{\textbf{LLM Engine}} & \multirow{2}{*}{\textbf{Method}} & \multicolumn{2}{c|}{\textbf{VQA-RAD}} & \multicolumn{2}{c}{\textbf{SLAKE}} \\
        \cmidrule{3-6}
        & & \textbf{Open} & \textbf{Closed} & \textbf{Open} & \textbf{Closed} \\
        \midrule
        \multirow{2}{*}\textbf{{DeepSeek-R1-Distill-Qwen-32B}} & \textbf{Med-InstructBLIP} & 32.41 & 61.76 & 42.82 & 59.38 \\
         & \cellcolor{mygray} \hspace{4pt}\textbf{+ \shortname{}}  & \cellcolor{mygray}35.81 {\tiny\textcolor[HTML]{AF6673}{(+3.40)}} & \cellcolor{mygray}67.28 {\tiny\textcolor[HTML]{AF6673}{(+5.52)}} & \cellcolor{mygray}43.87 {\tiny\textcolor[HTML]{AF6673}{(+1.05)}} & \cellcolor{mygray}70.91 {\tiny\textcolor[HTML]{AF6673}{(+11.53)}} \\
        \midrule
        \multirow{2}{*}\textbf{{DeepSeek-R1-Distill-Llama-70B}} & \textbf{Med-InstructBLIP} & 32.41 & 61.76 & 42.82 & 59.38 \\
         & \cellcolor{mygray} \hspace{4pt}\textbf{+ \shortname{}} & \cellcolor{mygray}34.28 {\tiny\textcolor[HTML]{AF6673}{(+1.87)}} & \cellcolor{mygray}66.18 {\tiny\textcolor[HTML]{AF6673}{(+4.42)}} & \cellcolor{mygray}44.34 {\tiny\textcolor[HTML]{AF6673}{(+1.52)}} & \cellcolor{mygray}70.43 {\tiny\textcolor[HTML]{AF6673}{(+11.05)}} \\
        \bottomrule
    \end{tabular}
    \caption{Performance of different open source LLMs as reasoning engine on VQA-RAD and SLAKE datasets.}
    \label{tab:results}
\end{table*}

\begin{table}[t!]
    \centering
    \tiny
    \renewcommand{\arraystretch}{1.2}
    \setlength{\tabcolsep}{6pt}
    \begin{tabular}{l c c| c c}
        \toprule
        \multirow{2}{*}{\textbf{Method}} & \multicolumn{2}{c|}{\textbf{VQA-RAD}} & \multicolumn{2}{c}{\textbf{SLAKE}} \\
        \cmidrule{2-5}
        & \textbf{Open} & \textbf{Closed} & \textbf{Open} & \textbf{Closed} \\
        \midrule
        SIRI~\cite{wang2023towards} & - & 45.80 & - & - \\
        KG-RAG~\cite{soman2024biomedical} & 35.56 & 52.57 & 46.71 & 66.34 \\
        BiomedGPT-S~\cite{zhang2023biomedgpt} & 13.40 & 57.80 & 66.50 & 73.40 \\
        AMANDA & 42.19 & 61.03 & 54.39 & 70.43 \\
        \bottomrule
    \end{tabular}
    \caption{Comparison of different methods on VQA-RAD and SLAKE datasets.}
    \label{tab:more_baseline}
\end{table}

\begin{table}[t!]
    \centering
    \tiny
    \renewcommand{\arraystretch}{1.2}
    \setlength{\tabcolsep}{10pt}
    \begin{tabular}{l c c| c c}
        \toprule
        \multirow{2}{*}{\textbf{Metric}} & \multicolumn{2}{c|}{\textbf{VQA-RAD}} & \multicolumn{2}{c}{\textbf{SLAKE}} \\
        \cmidrule{2-5}
        & \textbf{Open} & \textbf{Closed} & \textbf{Open} & \textbf{Closed} \\
        \midrule
        Average & 42.80 & 61.32 & 54.12 & 70.28 \\
        Std & 0.79 & 0.88 & 0.82 & 0.47 \\
        CV & 0.02 & 0.01 & 0.02 & 0.01 \\
        \bottomrule
    \end{tabular}
\caption{Stability analysis of \shortname{} across 5 runs with different random seeds. Std represents standard error and CV denotes coefficient of variation}
\label{tab:stability}
\end{table}


\section{Evaluation Metrics} 
For the closed-ended questions, we report the accuracy in a more strict way compared to prior work~\cite{li2024llava}. Instead of checking whether the ground-truth answer appears anywhere in the generated response, we only consider the first occurring yes/no-type word as the final prediction. This eliminates the inflated accuracy caused by long generated texts that include both "yes" and "no". For open-ended questions, we use recall to evaluate the ratio of ground-truth tokens that appear in the generated sequences. Different from the literature that selects from a fixed set of training answers, we do not provide any constraints on the model's open-ended responses. This makes our formulation closer to real open-ended questions but is intrinsically more challenging. For a fair comparison, we use the same strict accuracy metric for all methods. While this might lead to lower absolute numbers compared to what is typically reported, we believe it better reflects the true performance and is more meaningful.

\section{Additional Results of \shortname{} Framework on General MLLMs} \label{appendix: general-mllm}
While our main experiments demonstrate the effectiveness of \shortname{} on medical-specialized MLLMs, we further evaluate its generalization capability on general-domain MLLMs that lack medical pre-training. As shown in Table~\ref{tab: general-mllm}, our framework demonstrates strong generalization capability across different models. Specifically, when applied to InstructBLIP~\cite{instructblip}, \shortname{} achieves an average improvement of \textbf{14.68\%} over direct inference. These results suggest that our framework can effectively bridge the domain gap and enable general-purpose MLLMs to perform reliable medical visual reasoning.


\section{Compatibility with Different LLM Engines}
To demonstrate the versatility of \shortname{}, we evaluate its performance using different open-source LLMs as reasoning engines. As shown in Table~\ref{tab:results}, we test our framework with DeepSeek-R1-Distill-Qwen-32B and DeepSeek-R1-Distill-Llama-70B~\cite{guo2025deepseek} on the VQA-RAD and SLAKE datasets. When integrated with Med-InstructBLIP, both models show substantial improvements across all question types. Notably, with DeepSeek-R1-Distill-Qwen-32B, we achieve significant gains on closed-ended questions (+5.52\% on VQA-RAD, +11.53\% on SLAKE), while maintaining competitive performance on open-ended questions. Similar improvements are observed with DeepSeek-R1-Distill-Llama-70B, demonstrating that \shortname{} can effectively enhance medical visual reasoning capabilities regardless of the underlying LLM engine. These results indicate that our framework provides a cost-effective solution for improving Med-VQA performance without requiring specialized training or extensive computational resources.

\subsection{Comparison with Strong Baselines}
To provide a more comprehensive evaluation, we compare \shortname{} with several strong baselines, including both zero-shot and supervised approaches. The results in Table~\ref{tab:more_baseline} demonstrate \shortname{}'s effectiveness across different evaluation settings. Our framework significantly outperforms other zero-shot approaches, including SIRI~\cite{wang2023towards} (a multi-agent framework) and KG-RAG~\cite{soman2024biomedical} (which combines knowledge retrieval with LLM reasoning). Notably, \shortname{} achieves superior performance on VQA-RAD and competitive results on SLAKE compared to BiomedGPT-S~\cite{zhang2023biomedgpt}, despite the latter's advantage of supervised training on downstream tasks. These comprehensive comparisons validate the effectiveness of our training-free approach in medical visual reasoning tasks.

\subsection{Framework Stability}
We have thoroughly evaluated our framework's stability. As shown in Table~\ref{tab:stability}, we have conducted additional experiments, running LLaVA-Med v1.5 with \shortname{} 5 times with different seeds on different benchmarks. These results demonstrate the high stability of our framework, with standard deviations consistently below 1\% and coefficients of variation as low as 0.01-0.02. To put these variations in perspective, they are significantly smaller than the performance improvements our framework achieves over the baseline (e.g., an 8-25\% absolute improvement), confirming that \shortname{} provides stable and reliable enhancements across different medical visual reasoning tasks and models.

\section{Pseudo-Code of \shortname{} Framework} \label{appendix: pseudo-code}

\begin{algorithm}[t!]
\caption{\shortname{} Framework Pipeline}
\begin{lstlisting}[style=pythonstyle, basicstyle=\footnotesize\ttfamily]
def AMANDA(I: Image, Q: str) -> str:
    """
    Data-efficient Med-VQA
    Args:
        I: Input medical image
        Q: Input question
    Returns:
        Final answer
    """
    # Initialize reasoning history
    H = []
    
    # Initial Visual Understanding
    C, A_0 = Perceiver(I, Q)  # Generate medical caption and initial answer
    H.append((C, A_0))
    A_0 = Reasoner(Q, H)  # Initial reasoning
    confidence = Evaluator(A_0, H)
    
    # Medical Knowledge Augmentation
    while confidence < THRESHOLD:
        # Intrinsive Med-KA
        Q_sub, A_sub = Explorer(Q, H) 
        H.append((Q_sub, A_sub))
        
        # Extrinsive Med-KA
        K = Retriever(H)  
        H.append(K)
        
        # Re-reasoning with Enhanced Knowledge
        A_t = Reasoner(Q, H) 
        confidence = Evaluator(A_t, H)
    
    return A_t
\end{lstlisting}
\end{algorithm}

The algorithm illustrates how our framework orchestrates multiple specialized agents for collaborative medical reasoning. The process operates as follows:
\begin{itemize}[leftmargin=*, itemsep=0.0pt, topsep=0pt]
    \item The \texttt{Perceiver} agent first analyzes the medical image and generates a detailed caption along with an initial answer, establishing a foundation for visual understanding.
    
    \item The \texttt{Reasoner} agent then processes this initial information to generate a preliminary medical analysis based on the visual findings.
    
    \item The \texttt{Evaluator} agent assesses the confidence of the current answer by analyzing its consistency with the accumulated evidence.
    
    \item When confidence is insufficient, the \texttt{Explorer} agent generates strategic follow-up questions to probe deeper into critical visual details, while the \textit{Retriever} agent supplements the analysis with relevant medical knowledge from external sources.
    
    \item This iterative process continues until the \texttt{Evaluator} determines that sufficient confidence has been achieved, ensuring both comprehensive analysis and reliable diagnosis.
\end{itemize}

\section{Case Study} \label{appendix: case}
As shown in Table~\ref{tab:case_study}, this case study demonstrates how our \shortname{} framework effectively corrects initial misdiagnosis through comprehensive medical knowledge augmentation. Initially, the Med-MLLM baseline incorrectly identifies a rightward mediastinal shift. Our framework then initiates a systematic analysis through three key components. First, the \texttt{Perceiver} generates a detailed medical caption, establishing a foundation for understanding the image's key features. Second, through intrinsic Med-KA, the \texttt{Explorer} generates strategically designed sub-questions that progressively examine the mediastinal position from different perspectives. Third, the \texttt{Retriever} provides relevant medical knowledge about the mediastinum. Based on this augmented knowledge, the \texttt{Reasoner} generates a refined answer, which the \texttt{Evaluator} determines to be highly confident, demonstrating the effectiveness of our adaptive reasoning refinement. This case study exemplifies how our framework enhances medical reasoning through structured analysis and knowledge augmentation, leading to efficient and accurate diagnosis.

\section{Prompt Design of \shortname{}} \label{appendix: prompt}
In this section, we provide detailed prompt designs for each agent in our framework. First, Table~\ref{tab: perceiver_prompts} presents the caption generation prompts used by the \texttt{Perceiver} agent for comprehensive medical image understanding. Then, Table~\ref{tab: explorer_instructions} shows the structured instructions for the \texttt{Explorer} agent to generate coarse-to-fine follow-up questions. Table~\ref{tab:evaluator_instructions} details how the \texttt{Evaluator} agent assesses reasoning confidence. Finally, Tables~\ref{tab:closed_ended_reasoner_instructions} and \ref{tab:open_ended_reasoner_instructions} provide the \texttt{Reasoner} agent's prompts for handling closed-ended and open-ended questions respectively, demonstrating our framework's ability to adapt to different question types.

\begin{table*}[htbp]
\centering
\caption{A case study \shortname{} showing how our framework progressively refines the answer through knowledge augmentation.}
\setlength{\arrayrulewidth}{0.2mm}
\definecolor{mygray}{gray}{0.93}
\begin{tabular}{|>{\columncolor{mygray}}p{0.95\linewidth}|}
\hline
\textbf{Main Question:} Has the midline of the mediastinum shifted? \\
\\
\textbf{Medical Caption:} \\
This is a chest X-ray, which is a type of medical imaging study that uses low-energy X-rays to visualize the structures within the chest, including the lungs, heart, and bones. The chest X-ray shows bilateral ground-glass opacities in both lungs, which can be indicative of various lung conditions such as infections, inflammation, or other abnormalities. \\
\\
\textbf{Initial Answer:} Yes, the midline of the mediastinum has shifted to the right. \\

\textbf{Evaluator:} Confident Score : 1 \\
\\
\textbf{Coarse-to-fine question decomposition:} \\
Q1: Are there any visible signs of mediastinal shift, such as displacement of the trachea or heart? \\
A1: No, the chest X-ray does not show any visible signs of mediastinal shift, such as displacement of the trachea or heart. \\
\\
Q2: Is the position of the heart and trachea symmetrical and centered within the thoracic cavity? \\
A2: Yes, the position of the heart and trachea appears to be symmetrical and centered within the thoracic cavity. \\
\\
Q3: Are there any abnormalities in the lung volumes or pleural spaces that could contribute to a shift in the mediastinum? \\
A3: No, there are no abnormalities in the lung volumes or pleural spaces that could contribute to a shift in the mediastinum. \\
\\
\textbf{Retrieved Medical Knowledge:} \\
Lung disease Localizes in: cavity, chest, diaphragm, mediastinum. \\
\\
\textbf{Iteration 1 Answer:} No, the midline of the mediastinum has not shifted. \\
\textbf{Evaluator:} Confident Score : 4 \\\\

\textbf{Final Answer:} No, the midline of the mediastinum has not shifted. \\
\textbf{Ground Truth:} No. \\
\hline
\end{tabular}
\label{tab:case_study}
\end{table*}

\begin{table*}[htbp]
    \centering
    \caption{Prompts for the \texttt{Perceiver} Agent. Following LLaVA-Med's instruction tuning strategy, we randomly select one of these prompts to guide the \texttt{Perceiver} agent in generating comprehensive medical image descriptions.}
    \setlength{\arrayrulewidth}{0.2mm}
    \definecolor{mygray}{gray}{0.93}
    \begin{tabular}{|>{\columncolor{mygray}}p{0.95\linewidth}|}
    \hline
    \vspace{0.06cm}
    \textbf{PERCEIVER\_CAPTION\_PROMPTS:} \\
\vspace{0.06cm}
    \begin{itemize}
        \item Describe the following image in detail
        \item Provide a detailed description of the given image
        \item Give an elaborate explanation of the image you see
        \item Share a comprehensive rundown of the presented image
        \item Offer a thorough analysis of the image
        \item Explain the various aspects of the image before you
        \item Clarify the contents of the displayed image with great detail
        \item Characterize the image using a well-detailed description
        \item Break down the elements of the image in a detailed manner
        \item Walk through the important details of the image
        \item Portray the image with a rich, descriptive narrative
        \item Narrate the contents of the image with precision
        \item Analyze the image in a comprehensive and detailed manner
        \item Illustrate the image through a descriptive explanation
        \item Examine the image closely and share its details
        \item Write an exhaustive depiction of the given image
    \end{itemize}
\\
    \hline
    \end{tabular}
    \label{tab: perceiver_prompts}
\end{table*}

\begin{table*}[htbp]
    \centering
    \caption{\texttt{Explorer} agent instructions for generating follow-up questions.}
    \setlength{\arrayrulewidth}{0.3mm}
    \definecolor{mygray}{gray}{0.93}
    \begin{tabular}{|>{\columncolor{mygray}}p{0.95\linewidth}|}
    \hline
    \vspace{0.06cm}
    \textbf{EXPLORER\_SYSTEM\_PROMPT:} \\
    \vspace{0.06cm}
    You are an AI language model tasked with helping clinicians analyze medical images. Your goal is to decompose a primary clinical question into several sub-questions. By answering these sub-questions, it will be easier to arrive at a comprehensive answer for the main question. \\
    
    \textbf{Instruction:} Given a general caption that might not be entirely precise but provides an overall description, and a clinical question, generate a series of sub-questions to help thoroughly answer the main question. These sub-questions should guide the analysis step by step, focusing on the different aspects that could influence the final answer, keeping in mind clinical relevance and imaging characteristics. \\

    \textbf{Rules:}
    \begin{itemize}
        \item Break down the question into smaller parts following this hierarchical approach:
        \begin{itemize}
            \item[(a)] First, ask about general/overall observations 
            \item[(b)] Then, focus on specific anatomical regions or structures
            \item[(c)] Finally, ask about detailed findings or specific characteristics
        \end{itemize}
        \item Consider these aspects in your questions:
        \begin{itemize}
            \item Presence or absence of specific findings
            \item Characteristics of structures (e.g., size, shape, alignment)
            \item Orientation and positioning of the patient or organs
            \item Comparison of abnormal vs. normal findings
        \end{itemize}
        \item The number of sub-questions should be less or equal to \{max\_sub\_questions\}.
        \item Order your questions from general to specific (coarse to fine-grained).
    \end{itemize}

    \textbf{Format:} \\
    Sub-question 1: [General observation question] \\
    Sub-question 2: [Specific anatomical region question] \\
    Sub-question 3: [Detailed finding question] \\
    ... \\
    \\

    \textbf{EXPLORER\_PROMPT:} \\
    Image description: \{caption\} \\
    Main question: \{question\} \\
    History: \{history\} \\

    Please generate a series of follow-up questions following a coarse-to-fine approach. Start with general observations and progressively move to more specific details. \\

    \hline
    \end{tabular}
    \label{tab: explorer_instructions}
\end{table*}

\begin{table*}[htbp]
    \centering
    \caption{Open-ended \texttt{Reasoner} instructions.}
    \setlength{\arrayrulewidth}{0.2mm}
    \definecolor{mygray}{gray}{0.93}
    \begin{tabular}{|>{\columncolor{mygray}}p{0.95\linewidth}|}
    \hline
    \vspace{0.06cm}
    \textbf{OPEN\_ENDED\_REASONER\_SYSTEM\_PROMPT:} \\
    \vspace{0.06cm}
    You are a medical AI assistant with rich visual commonsense knowledge and strong reasoning abilities. \\
    
    \textbf{You will be provided with:}
    \begin{enumerate}
        \item A main question about an image.
        \item An imperfect initial answer to the main question provided by a visual AI model. Note that the answers may not be entirely precise.
        \item A general caption that might not be entirely precise but provides an overall description.
        \item Some conversation history containing follow-up questions and answers.
        \item Some grounded medical information.
        \item Some similar examples with their answers for reference.
    \end{enumerate}

    \textbf{Your goal:} Based on the above information, find the answer to the main question. \\

    \textbf{Rules:}
    \begin{enumerate}
        \item Begin with a brief paragraph demonstrating your reasoning and inference process. Start with the format: "Analysis:".
        \item Be logical and consistent in evaluating all clues, including as many relevant details as possible.
        \item Use similar examples to inform your reasoning.
    \end{enumerate}

    \textbf{Response Format:}
    \begin{verbatim}
    Analysis: xxxxxx.

    Answer: xxxxxx
    \end{verbatim}
    \\

    \textbf{OPEN\_ENDED\_REASONER\_PROMPT:} \\
    Imperfect image description: \{caption\} \\
    Open-ended question: \{question\} \\
    Initial answer: \{initial\_answer\} \\
    History: \\
    \{history\} \\
    Additional information: \{rag\_context\} \\

    Please provide a detailed answer to the open-ended question based on all the information provided. \\

    \hline
    \end{tabular}
    \label{tab:open_ended_reasoner_instructions}
\end{table*}

\begin{table*}[htbp]
    \centering
    \caption{Closed-ended \texttt{Reasoner} instructions.}
    \setlength{\arrayrulewidth}{0.2mm}
    \definecolor{mygray}{gray}{0.93}
    \begin{tabular}{|>{\columncolor{mygray}}p{0.95\linewidth}|}
    \hline
    \vspace{0.06cm}
    \textbf{CLOSED\_ENDED\_REASONER\_SYSTEM\_PROMPT:} \\
    \vspace{0.06cm}
    You are a medical AI assistant with rich visual commonsense knowledge and strong reasoning abilities. \\

    \textbf{You will be provided with:}
    \begin{enumerate}
        \item A main question about an image.
        \item An imperfect initial answer to the main question provided by a visual AI model. Note that the answers may not be entirely precise.
        \item A general caption that might not be entirely precise but provides an overall description.
        \item Some conversation history containing follow-up questions and answers.
        \item Some grounded medical information.
        \item Some similar examples with their answers for reference.
    \end{enumerate}

    \textbf{Your goal:} Based on the above information, find the answer to the main question. \\

    \textbf{Rules:}
    \begin{enumerate}
        \item Begin with a brief paragraph demonstrating your reasoning and inference process. Start with the format: "Analysis:".
        \item Be logical and consistent in evaluating all clues, but aim to preserve the initial answer unless strong contradictions arise.
        \item Use similar examples to inform your reasoning.
    \end{enumerate}

    \textbf{Response Format:}
    \begin{verbatim}
    Analysis: xxxxxx.

    Answer: [Yes/No] or [Selected Option]
    \end{verbatim}
    \\

    \textbf{CLOSED\_ENDED\_REASONER\_PROMPT:} \\
    Imperfect image description: \{caption\} \\
    Closed-ended question: \{question\} \\
    Initial answer: \{initial\_answer\} \\
    History: \\
    \{history\} \\
    Additional information: \{rag\_context\} \\

    Please provide an answer to the closed-ended question based on all the information provided. \\

    \hline
    \end{tabular}
    \label{tab:closed_ended_reasoner_instructions}
\end{table*}

\begin{table*}[ht]
    \centering
    \caption{\texttt{Evaluator} agent instructions for assessing confidence levels in medical image analysis responses.}
    \setlength{\arrayrulewidth}{0.2mm}
    \definecolor{mygray}{gray}{0.93}
    \begin{tabular}{|>{\columncolor{mygray}}p{0.95\linewidth}|}
    \hline
    \vspace{0.06cm}
    \textbf{EVALUATOR\_SYSTEM\_PROMPT:} \\
    \vspace{0.06cm}
    You are a medical AI assistant specialized in evaluating answers for medical image analysis. \\

    \textbf{You will be provided with:}
    \begin{enumerate}
        \item A main question about a medical image.
        \item A general caption that might not be entirely precise and may contain false information.
        \item Current answer.
        \item History of the conversation.
        \item Examples from in-context learning.
    \end{enumerate}

    \textbf{Your goal:}
    \begin{enumerate}
        \item Assess the confidence level of a given answer and provide a brief explanation.
        \item Provide a confidence score from 1 to 5, where 1 means completely uncertain and 5 means very certain.
        \item Use examples from in-context learning to assist in evaluating the answer.
    \end{enumerate}

    \textbf{Evaluation Criteria:}
    \begin{itemize}
        \item \textbf{Contradictory Evidence}: Look for any information that strongly contradicts the current answer. If significant conflicting information is found, reduce the confidence level.
    \end{itemize}

    \textbf{Scoring Guidance:}
    \begin{itemize}
        \item \textbf{Score 5}: The answer is accurate, consistent with all provided information, and has no significant conflicting evidence.
        \item \textbf{Score 4}: The answer is mostly correct, with minor issues or slight uncertainty.
        \item \textbf{Score 3}: The answer is generally acceptable, with some uncertainty or minor inconsistencies, but it mostly aligns with the question.
        \item \textbf{Score 2}: The answer has notable inaccuracies or lacks consistency, with some conflicting information present.
        \item \textbf{Score 1}: The answer is largely incorrect, inconsistent, or contains major contradictions with the provided information.
    \end{itemize}

    \textbf{Response Format:}
    \begin{verbatim}
    Score: [1-5]
    Explanation: [Your explanation]
    \end{verbatim}
    \\

    \textbf{EVALUATOR\_PROMPT:} \\
    Imperfect image description: \{caption\} \\
    Main question: \{question\} \\
    Current answer: \{answer\} \\
    History: \\
    \{history\} \\

    Please evaluate the confidence level of the current answer and provide a brief explanation. \\

    \hline
    \end{tabular}
    \label{tab:evaluator_instructions}
\end{table*}